%% file: main.tex
\documentclass[runningheads]{llncs}

\usepackage{eccv}

\usepackage{eccvabbrv}
\usepackage{xr}
\usepackage{hyperref}
\usepackage{xr-hyper} 

\usepackage{sidecap}
\usepackage{floatrow}
\usepackage{wrapfig,booktabs}
\usepackage[ruled,vlined]{algorithm2e}
\usepackage{enumitem}
\usepackage{graphicx} 
\usepackage{amsfonts}
\usepackage{amsmath}
\DeclareMathOperator*{\argmin}{arg\,min}
\usepackage{xcolor}
\usepackage{gensymb}
\usepackage{array}
\usepackage{bm}
\usepackage{tabularx}
\usepackage{adjustbox}
\newcolumntype{Y}{>{\centering\arraybackslash}X}
\usepackage[accsupp]{axessibility}  

\usepackage{hyperref}

\usepackage{orcidlink}

\include{notation_3Dtree}
\begin{document}


\title{TreeSRNF: Square-Root Normal Fields for Generative Modelling of the Geometric and Structural Variability in Tree-like 3D Objects}

\titlerunning{TreeSRNF: Elastic Shape Analysis of Tree-like 3D Shapes}

\author{Tahmina Khanam\inst{1}\orcidlink{0000-0002-9549-2700} \and
Hamid Laga\inst{1}\orcidlink{0000-0002-4758-7510} \and
Mohammed Bennamoun\inst{2}\orcidlink{0000-0002-6603-3257} \and
Guanjin Wang\inst{1}\orcidlink{0000-0002-5258-0532} \and
Ferdous Sohel\inst{1}\orcidlink{0000-0003-1557-4907} \and
Farid Boussaid\inst{2}\orcidlink{0000-0001-7250-7407} \and
Anuj Srivastava\inst{3}\orcidlink{0000-0001-7406-0338}}

\authorrunning{F.~Author et al.}

\institute{Murdoch University, WA, Australia. 
\email{34719017@student.murdoch.edu.au, \{H.Laga, Guanjin.Wang,F.Sohel\}@murdoch.edu.au} \and
University of Western Australia, WA, Australia.  \email{\{mohammed.bennamoun, farid.boussaid\}@uwa.edu.au} \and Johns Hopkins University, USA.  \email{anuj.srivastava@jhu.edu}.}

\maketitle

\begin{abstract}
We introduce a novel mathematical framework for analyzing and generating complex tree-shaped 3D objects,  such as botanical trees and plants, which deform both in their 3D geometry and branching structure. Unlike previous works, which either consider only the skeletal structure of tree-like objects or approximate their 3D geometry using branch thickness, the proposed framework accurately models both the 3D geometry of the tree branches and the way they are interconnected. In this paper, we first generalize the Square Root Normal Fields (SRNF) representation, originally proposed for the statistical analysis of genus-0 surfaces, to tree-shaped 3D objects. We then treat tree-shaped 3D objects as points on a novel Riemannian tree-shape space equipped with a novel Riemannian metric that measures the amount of surface bending and stretching, and structural changes one needs to apply to one 3D tree-shape to align it with another. This way, deformations
become trajectories in this novel tree-shape space. We analyze the theoretical properties of this novel tree-shape space and the corresponding metric and develop algorithms for computing point-wise and branch-wise correspondences and geodesic paths between complex 3D trees. We finally show how to use these building blocks for \textbf{(1)} computing statistical summaries, \ie means and modes of variation, of collections of tree-shaped 3D objects, and \textbf{(2)} synthesizing novel tree-shaped 3D objects by sampling from probability distributions fitted to a population of tree-shaped  3D objects. We demonstrate the performance and utility of the proposed framework on real and synthetic plants and botanical trees and show that it significantly outperforms the state-of-the-art. Additional results and source code are available at \url{https://tahmina979.github.io/Tree_in_SRNF/}
\end{abstract}

\section{Introduction}
\label{sec:introduction}
We introduce a novel mathematical framework and a set of algorithms for modelling and statistically analyzing the geometric and structural variability in tree-like 3D objects. State-of-the-art statistical shape analysis methods~\cite{jermyn2017elastic,laga2017modeling,laga2017numerical,laga20224d} are mostly limited to 3D objects of fixed branching structure, \ie objects that bend and stretch while their branching structure remains static. In nature, however, a large number of biological objects, from blood vessels and airway trees in the human body to the roots and shoots of botanical trees and plants, are tree-shaped. Both their 3D geometry and branching structure, which are tightly linked to their functionality,  deform as they grow, interact with their surrounding environment, and compete for resources.
Understanding and mathematically modelling such variability is important in many applications, from plant biology to neuroscience and health. 

Unlike objects with fixed topology, tree-like 3D objects exhibit variability in \textbf{(1)} their geometry, in terms of the 3D shape of individual branches, which can bend and stretch, and \textbf{(2)} the structural relationships between those branches as new branches emerge and others die over time.  Existing methods that attempted to address these challenges define a tree-shape space where tree-shapes can be seen as points in this space while paths in this space correspond to the deformation of one tree-shape onto another. By equipping the tree-shape space with a proper metric that quantifies geometric and topological variations, one can develop algorithms for computing registrations, \ie one-to-one branchwise and pointwise correspondences, and geodesic paths, \ie shortest deformation paths with respect to the metric, between trees. Early methods that follow this paradigm~\cite{billera2001geometry,owen2010fast} only model the tree topology, completely ignoring the geometry of its branches. These methods have been extended first by  Feragen \etal~\cite{feragen2010geometries,feragen2011means,feragen2012toward,feragen2013tree} and Duncan \etal~\cite{duncan2018statistical} to take into account the shape of the skeletal curves of the branches, and later by   Wang \etal~\cite{wang2018shape,wang2018statistical,wang2023elastic} by augmenting skeletal curves with their local thickness. This representation, which is an over-simplification of the 3D geometry of branches with local cylinders,  fails to capture the detailed and accurate geometry of branches. Thus, these methods fail to accurately model the geometric and structural variability of tree-shaped 3D objects.

In this paper, we propose the first comprehensive framework that models both the full 3D geometric and structural variability in tree-shaped 3D objects. 
Our key contribution is the generalization of the Square Root Normal Field (SRNF), originally proposed for the analysis of genus-0 surfaces~\cite{jermyn2012elastic,laga2017numerical}, to tree-like 3D shapes that bend and stretch, and change their branching structure. 

This novel representation results in a new tree-shape space, which we refer to as TreeSRNF, equipped with a metric that measures surface bending and stretching, as well as topological changes in the form of branch sliding (\cref{sec:representation}). In this paper, we establish the theoretical framework and analyze the properties of this tree-shape space and the associated metric (\cref{sec:elastic_metric}). We will then develop tools for computing point-wise and branch-wise correspondences (\cref{sec:registration}) and geodesic deformations (\cref{subsec:geod}) between complex tree-like 3D objects, even in the presence of large bending, stretching, and topological variation.
We will also show how to use these building blocks to  \textbf{(1)} compute statistical summaries, \eg average 3D shape and modes of variation,  and mathematically model the variability in a population of tree-like 3D objects by characterizing the population with probability distributions, and \textbf{(2)} generate novel tree-like 3D objects by sampling from the learned probability distributions (\cref{subsec:summary_stat}).  
We demonstrate and evaluate the performance of the proposed framework on real and synthetic tree-like 3D shapes such as botanical trees and plants (\cref{sec:results}) and show that it outperforms the state-of-the-art in terms of representation quality and accuracy.

\section{Related work}
\label{sec:related_work}
Statistical 3D shape analysis is an extensively studied problem, with key subproblems including registration, geodesics computation, and statistical summarization. Early contributions focused on 3D objects that bend and stretch while preserving their topology.  In these works, the shape of a 3D object is treated as a point in a high-dimensional shape space equipped with an appropriate metric that quantifies the amount of bending and stretching required to align one shape to another. This way, both registration and geodesics computation between two 3D shapes can be jointly formulated as the problem of finding the shortest path between their corresponding points in the shape space.  Existing works differ in the nature of the shape space and the metric they use~\cite{laga2018survey}.

Early works employ the $\ltwo$ metric, which is only suitable for rigid shapes or 3D shapes that undergo small non-rigid deformations. To handle large non-rigid deformations, Kilian \etal~\cite{kilian2007geometric}  quantify shape differences using deviations from rigidity and isometry. Other works measure bending using differences in the shape operator~\cite{zhang2015shell,wirth2011continuum} or the second fundamental form~\cite{heeren2012time}. Stretching, on the other hand, is measured using the Cauchy-Green strain tensor~\cite{zhang2015shell,wirth2011continuum} or differences in the first fundamental form~\cite{heeren2012time}. These physics-motivated metrics are computationally very expensive, as registration and statistical summarization require solving complex nonlinear optimization problems. Addressing the computational cost, Jermyn \etal~\cite{jermyn2012elastic, laga2017numerical} introduced the  Square Root Normal Field (SRNF) representation, where the $\ltwo$ metric approximates a partial elastic metric, significantly reducing the computational complexity. These representations, however, do not capture topological changes observed in tree-like 3D shapes.

Unlike manifold surfaces, tree-like 3D shapes not only bend and stretch but also exhibit topological variation in their branching structure. Addressing such structural variability, Billera \etal~\cite{billera2001geometry} and Owen \etal~\cite{owen2010fast} introduced  the concept of continuous tree spaces, which model changes in topology. The representation, however, ignores the 3D geometry of tree branches. Recent methods that consider tree shapes are primarily focused on the reconstruction~\cite{Liu_2025_CVPR,wang2025autoregressive,lee2024tree,zhou2023deeptree,li2024interactive,zhou2025treestructor,li2025stressful} of static or dynamic 3D tree shapes, while others focused on the registration~\cite{zhang2023spatio,wang2022plantmove,lobefaro2023estimating,magistri2020segmentation,chebrolu2021registration,pan2021multi}. These methods do not offer metrics or algorithms for deriving statistical summaries of collections of 3D tree-like shapes. Feragen \etal~\cite{feragen2010geometries,feragen2011means,feragen2012toward,feragen2013tree} introduced the concept of tree-shape spaces and developed corresponding metrics to support statistical analysis of tree-like shapes. These  methods, which were  limited to trees with simple branching structures, have been extended by Wang \etal~\cite{wang2018shape,wang2018statistical} to complex trees by pre-computing the correspondences. However, when applied to shapes with significant structural differences, it results in excessive shrinkage along the geodesic path. Duncan \etal~\cite{duncan2018statistical} proposed an alternative metric  that measures topological changes in terms of branch sliding along parent branches. Khanam \etal~\cite{khanam2024riemannian} extended the representation to 4D tree-like shapes. The method, however,  is limited to skeletal representations. Guan \etal~\cite{wang2023elastic} extended this idea by approximating the 3D geometry of branches using their skeletal curves augmented with local thickness. This over-simplification fails to capture the detailed 3D geometry of trees. Unlike previous work, this paper proposes the first framework that allows the statistical analysis of the full 3D geometry and structure of tree-like 3D objects. 
\begin{figure}[t] 
    \centering
    \includegraphics[width=1\linewidth]{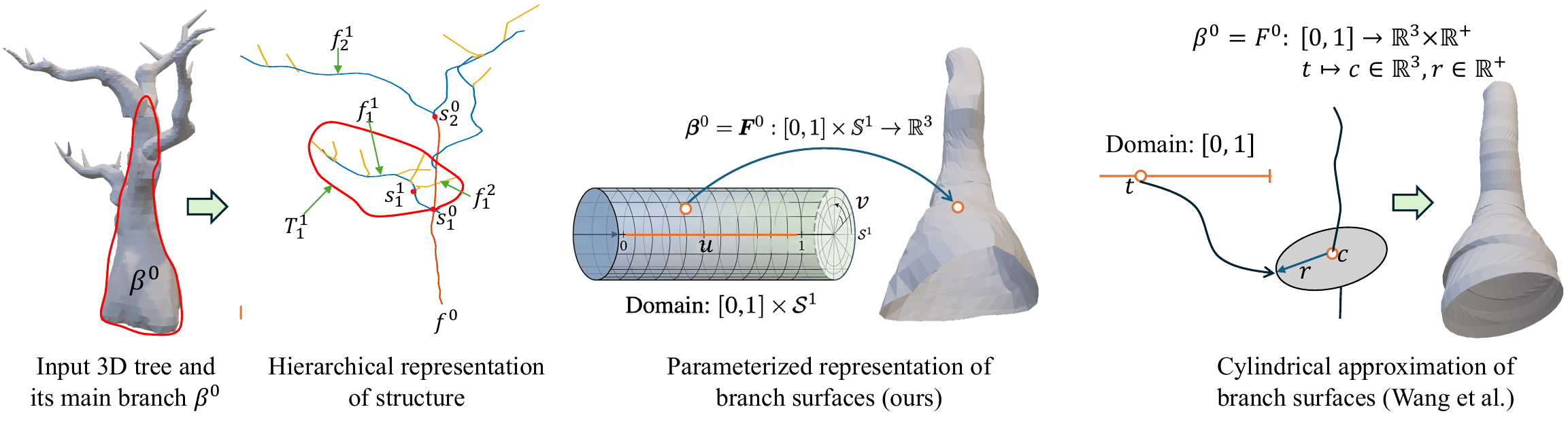}
    \caption{An illustration of the proposed representation, compared to Wang \etal~\cite{wang2023elastic}'s.}
    \label{fig:representation_tree}
\end{figure}

\section{Representation}
\label{sec:representation}
We represent a branch $\branch$ of a 3D tree structure $\threeDTree$ as a parameterised surface $\brachsurface$  of the form  $\brachsurface:\paramdomain \rightarrow \rthree$. Since, in general, the branches of tree-like objects are elongated surfaces, we define the parameterization domain $\paramdomain$ as an open cylinder of unit radius and height, \ie   $\paramdomain = \unitcylinder$.  Let $\parameters = (u,v) \in \paramdomain$ denote the cylindrical coordinates, where $u \in [0,1]$ is the height of the cylinder and $v \in [0,2\pi]$ is the circumference with radius of one. Let $\preshapebrachsurface$ denote the preshape space of such surfaces, after normalization for translation and scale.

Next, we construct the hierarchical representation of a 3D tree $\threeDTree$ by connecting each branch to its parent branch at the corresponding bifurcation point. We set the bifurcation point on the medial axis of branch $\branch$ which is parameterized as a function $\medialaxis: [0, 1] \rightarrow \rthree$. Let $\branch^0$ denote the main branch, also referred to as the first layer in the tree hierarchy, with $k^0$ subtrees $\{\threeDTree^1_i\}_{i=1}^{k^0}$ attached to it. The initial point of each subtree is attached to the medial axis of its parent branch $\branch^0$ at the corresponding bifurcation points $\bifurcation^0_i, i=1,\dots,k^0$. Thus, a one-layer 3D tree is defined as
$\threeDTree^0 = (\branch^0, \{\threeDTree^1_{i}, \bifurcation^0_{i}\}_{i=1}^{k^0})$. A 3D tree with $L$ layers, denoted by $\threeDTree^{L-1}$,  can be recursively defined  as  $\threeDTree^l = (\branch^l,\{\threeDTree^{l+1}_i,\bifurcation^{l}_i\}_{i=1}^{k^l}), \text{ for } l = 1, \dots, L-1$. Here,   $\branch^l$ is the main branch of  $\threeDTree^l$, $k^l$ is the number of subtrees  attached to the branch $\branch^l$, and $\bifurcation^{l}_i \in [0, 1]$ is a bifurcation point on the medial axis $\medialaxis^l$ of the branch $\branch^l$. \cref{fig:representation_tree} summarizes the representation. To ensure the representation is invariant to translation and scale, we first translate the entire tree so that the main branch's starting point is at the origin and scale it so that the entire tree fits within a unit sphere. We refer to the space of all such trees normalized for translation and scale as the \emph{pre-tree-shape space} and denote it by $\preshapespaceThreeDTreees$. Here, scaling is optional as it may not be desirable for some applications such as growth analysis. 

\section{Elastic metric for comparing 3D tree shapes}
\label{sec:elastic_metric}
A key component of statistical 3D shape analysis is a metric that measures shape differences, or distances between points in the tree-shape space. In our case, a proper metric should quantify \textbf{(1)} branch deformations, \ie the bending and stretching of branches (\cref{subsec:elastic_metric_for_branch}), and \textbf{(2)} changes in the branching structure of trees (\cref{sec:tree_metric}).
The metric will then be used in subsequent analysis tasks for registration, geodesics computation, and statistical summarization. 

\subsection{Elastic metric for comparing branches }
\label{subsec:elastic_metric_for_branch}
To compare the shape of two tree branches $\branch_1$ and $\branch_2$, we propose an elastic metric $\distance$ that measures the bending and stretching of the branches after shape-preserving transformations, \ie translations and reparameterizations, are factored out. Let $\reparmspace$ be the space of  reparameterizations $\reparm$ of the domain $\paramdomain$ to itself, \ie $\reparm: \paramdomain \to \paramdomain$ such that $\reparm$ is a diffeomorphism. Since the branches are already normalized for translation, the distance $\distance$  can be mathematically formulated as:
\begin{equation}
   \distance(\branch_1, \branch_2)= \min_{\reparm \in \reparmspace} \distancebranch\left(\brachsurface_1, \brachsurface_2 \circ \reparm \right).
\label{eq:distance_branches}
\end{equation}

\noi Here, $ \distancebranch$ measures the amount of bending and stretching one needs to apply to $\brachsurface_2 \circ \reparm$ in order to align it onto $\brachsurface_1$.
Bending can be quantified by measuring changes in the orientation of the normal vectors. Stretching, on the other hand, can be decomposed into two components: stretching resulting from changes in local surface area and stretching that preserves local surface area. While metrics exist for quantifying these physical deformations, they are highly non-linear, making solutions to the optimization problem of Eqn.~\eqref{eq:distance_branches}, and the subsequent registration and geodesics-computation problems time-consuming.

To address this issue, Jermyn \etal~\cite{jermyn2012elastic} introduced a representation termed Square-Root Normal Field (SRNF) and showed that the $\ltwo$ metric in the space of SRNFs is equivalent to a weighted sum of bending and stretching in the original space of surfaces. This significantly simplifies downstream tasks: instead of operating in the original space of surfaces, one can map the surfaces to their SRNF representations, perform all the analysis tasks in that space using the standard $\ltwo$ metric, and then map the results back to the original space of surfaces. Formally, the SRNF of a  surface $\brachsurface \in \preshapebrachsurface$  is defined as:
\begin{equation}
    \srnfbranch(\brachsurface)=\srnf, \,\text{such that} \, \srnf(u,v)=\frac{\surfacenormal(u,v)}{\sqrt{\norm{\surfacenormal(u,v)}}}\,.
\label{eq:srnf}
\end{equation}

\noi Here, $\surfacenormal=\frac{\partial\brachsurface}{\partial u}\times\frac{\partial\brachsurface}{\partial v}$ is the normal field to $\brachsurface$ at $(u,v) \in \paramdomain$. Let $\preshapespacesrnfbranch$ be the space of SRNFs. The SRNF representation has many properties that are important for statistical shape analysis. \textbf{First}, it is translation invariant. \textbf{Second}, the action of the reparameterization group on the SRNF is by isometry. In other words, $\forall \reparm \in \reparmspace$ and $\forall\brachsurface_1,\brachsurface_2\in\preshapebrachsurface, \norm{\srnf_1-\srnf_2}=\norm{(\srnf_1\circ\reparm)\sqrt{\Dot{\reparm}}-(\srnf_2\circ\reparm)\sqrt{\Dot{\reparm}}}$, where $\srnf_i=\srnfbranch(\brachsurface_i)$ for $i\in \{1,2\}$. For simplicity of notation, in the remainder of the paper, we denote $(\srnf\circ\reparm)\sqrt{\Dot{\reparm}}$ by $(\srnf,\reparm$). \textbf{Finally}, the  $\ltwo$ metric in SRNF space $\preshapespacesrnfbranch$ is analogous to the partial elastic metric in the original space of shapes $\preshapebrachsurface$.
We leverage this last property to redefine the metric of Eqn.~\eqref{eq:distance_branches} using the $\ltwo$ metric in the SRNF space. In practice, we further decompose the reparameterization $\reparm$ into a global reparameterization $\greparm \in \reparmspace$, which models the global rotation of the parameterization grid, and local reparameterization, which we also denote by $\reparm$ for simplicity. This way, Eqn.~\eqref{eq:distance_branches} becomes 
$
    \distance(\branch_1, \branch_2) =  \min_{\greparm,\reparm \in \reparmspace} \| \left(\srnf_1, (\srnf_2,\greparm,\reparm) \right)\|^2$. This formulation significantly reduces complexity and computation time, as downstream tasks can be solved in the SRNF space using the standard $\ltwo$ metric.

\subsection{Extension to 3D trees}
\label{sec:tree_metric}
A  complete 3D tree can now be  represented recursively as $\srnfthreeDTree^l = (\srnf^l,\{\srnfthreeDTree^{l+1}_i,s^{l}_i\}_{i=1}^{k^l})$,  for  $l = 1, \dots, L-1$ where $L$ is the total number of layers,  $\srnf^l$ is the SRNF of the main branch of  $\srnfthreeDTree^l$, $k^l$ is the number of subtrees attached to the $l$-th branch, and $\bifurcation^{l}_i \in [0, 1]$ is a bifurcation point on the medial axis of the $l$-th branch. Let $\preshapesrnf$ denote the pre-tree shape space of such trees defined in the SRNF space.

\begin{wrapfigure}{r}{0.6\textwidth}
    \centering
    \includegraphics[width=\linewidth]{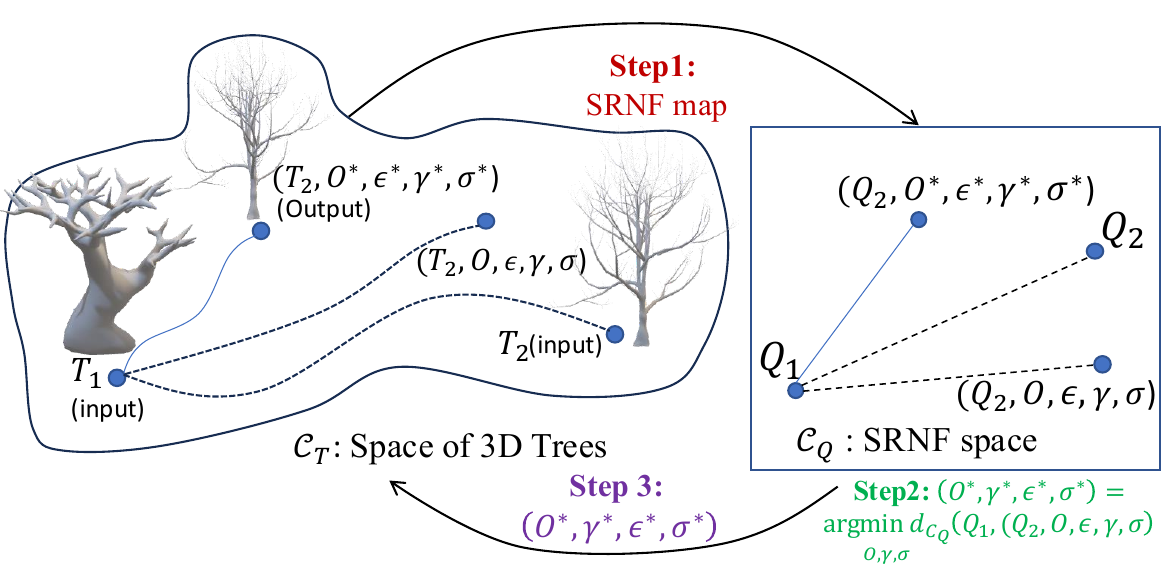}
    \caption{Overview of the proposed registration framework. We first map all trees into the SRNF space $\preshapesrnf$. We then solve~\cref{eq:optim_for_tree} to get optimal transformations  $(\rotation\optimal,\greparm\optimal,\reparm\optimal,\permutation\optimal)$. Finally, we apply these transformations to $\threeDTree_2$ to align it to $\threeDTree_1$.}
    \label{fig:spatial_registration} 
\end{wrapfigure}

A good metric that compares and measures distances between two complete tree shapes needs to also take into account branch-wise correspondences, differences in their branching structure, and also differences in the number of branches and different levels of the tree hierarchy. Following Duncan \etal~\cite{duncan2018statistical}, given the SRNFs  $\srnfthreeDTree_1$ and  $\srnfthreeDTree_2$ of two trees,  we define the  metric iteratively as:
\begin{equation}
     d(\srnfthreeDTree_1, \srnfthreeDTree_2) =\min_{\rotation\in\rotationspace, \greparm\in\reparmspace, \reparm\in \reparmspace,\permutation\in\permutationspace}d_{\preshapesrnf}\left(\srnfthreeDTree_1, (\srnfthreeDTree_2,\rotation,\greparm,\reparm,\permutation)\right).
     \label{eq:optim_for_tree}
\end{equation}

\noi Here, $\rotation\in\rotationspace$ is the global rotation, $\permutation\optimal \in \permutationspace$ is the permutation of the order of the tree branches,  $\permutationspace$ is the space of such permutations, and 
\begin{equation}
 	\label{eq:metric_for_trees}
  \small{
 	d_{\preshapesrnf}\left(\srnfthreeDTree_1, \srnfthreeDTree_2\right)  = \lambda_m \parallel \srnf^0_{1} -\srnf^0_{2}\parallel^2  + 				
 				\lambda_p \sum_{i=1}^{n} \left({\bifurcation}^{i}_{1} - {\bifurcation}^{i}_{2}\right)^2 
 + \lambda_s \sum_{i=1}^{n} d \left( \srnfthreeDTree_1^i, \srnfthreeDTree_2^{i} \right),
 }
 \end{equation}

\noi where $n\in \{k^0,\dots, k^l\}$ , $k^l$ is the number of subtrees attached to the branch $\srnf_1^l$ or $\srnf_2^l$, and $l = 1, \dots, L-1$ is the total number of layers. 
$s_1$ and $s_2$ are the corresponding bifurcation points on the parent branch.  The first term of ~\cref{eq:metric_for_trees} quantifies the bending and stretching of the main (parent) branch. The second term measures the bifurcation distance between subtrees to quantify the topological variation while the third term measures geometrical variation between the subtrees. The parameters $(\lambda_m,\lambda_p,\lambda_s)$ control the importance of each of these terms.  
Note that this formulation assumes that  $\srnf_1^l$ and $\srnf_2^l$ have the same number of branches. This, however, is not the case in practice. To handle trees of different numbers of branches, we add null branches so that both of the branches $\srnf_1^l$ or $\srnf_2^l$ have exactly the same $k^l$ subtrees. For a null branch, all the cylindrical coordinates in the domain $\paramdomain$ are mapped to the point $(0,0,0)\in\rthree$.

\section{Registration}
\label{sec:registration}
The problem of registering a source tree-shape onto a target tree-shape can be defined as that of finding the global optimal rotation $\rotation\optimal\in\rotationspace$ of the source tree-shape, global and local reparameterizations $\greparm\optimal,\reparm\optimal\in\reparmspace$ of each of its branches, and the optimal permutation $\permutation\optimal \in \permutationspace$ of the order of its branches in order to bring it as close as possible to the target tree-shape. Mathematically, this can be formulated as an optimization problem of the form:\begin{equation}
     (\rotation\optimal, \greparm\optimal,\reparm\optimal,\permutation\optimal)=\argmin_{\rotation\in\rotationspace, \greparm\in\reparmspace, \reparm\in \reparmspace,\permutation\in\permutationspace}d_{\preshapesrnf}\left(\srnfthreeDTree_1, (\srnfthreeDTree_2,\rotation,\greparm,\reparm,\permutation)\right),
     \label{eq:optim_for_tree}
\end{equation}

\noi where $d_{\preshapesrnf}$ is the metric defined in~\cref{eq:metric_for_trees}. 
The notation $ (\srnfthreeDTree_2,\rotation,\greparm,\reparm,\permutation)$  in~\cref{eq:optim_for_tree} refers to the process of applying the transformations $(\rotation,\greparm,\reparm,\permutation)$ to $\srnfthreeDTree_2$. 

To solve this optimization problem, we exploit the recursive nature of the metric $d_{\preshapesrnf}$ to derive an iterative algorithm. We start by computing the optimal global rotation $\rotation\optimal$ of the entire tree $\srnfthreeDTree_2$ by minimizing~\cref{eq:metric_for_trees} with respect to $\rotation$ while setting all the other parameters fixed. Let $\srnfthreeDTree_2=\rotation\optimal(\srnfthreeDTree_2)$. We then proceed in an iterative manner to find the remaining optimums by minimising:
\begin{equation}
 	\label{eq:tree_optim}
    \resizebox{0.93\hsize}{!}{$\displaystyle{
 	d_{\preshapesrnf}\left(\srnfthreeDTree_1, (\srnfthreeDTree_2,\greparm,\reparm,\permutation) \right)  = \lambda_m \parallel \srnf^0_{1} - (\srnf^0_{2},\greparm^0, \reparm^0) \parallel^2  + 				
 				\lambda_p \sum_{i=1}^{n} \left({\bifurcation}^{i}_{1} - {\bifurcation}^{\permutation(i)}_{2}\right)^2 
 + \lambda_s \sum_{i=1}^{n} d \left( \srnfthreeDTree_1^i, (\srnfthreeDTree_2^{\permutation(i)},\greparm^i,\reparm^i,\permutation(i)) \right).}
 	$}
 \end{equation}
 
\noi Here,  
    $\greparm^0$ is the global reparameterization  of the main branch of $\srnf^l$ and $\greparm^i$ is the global reparameterization of its subtrees and  
    $\reparm^0$ is the local reparameterization of the main (parent) branch of $\srnf^l$ and $\reparm^i$ is the local reparameterization of its subtrees. 
  To quantify the topological variation, 
    we define $\permutation^0$ as the relative optimal order of the parent branch of $\srnf^l$ on its corresponding parent and $\{\permutation^i\}_{i=1}^{k^l}$ is the permutation of the order of the subtrees attached to $\srnf^l$. 

A high-level overview of the iterative registration procedure between two 3D trees is summarized in~\cref{alg:registration} and~\cref{fig:spatial_registration}. We solve the problem of finding which side branch of $\srnfthreeDTree_1$ corresponds to which side
branch of $\srnfthreeDTree_2$ at any level by solving the last two terms of ~\cref{eq:tree_optim}, as a linear assignment problem. Let $n_1$ be the number of side branches (at any level) of the main branch of $\srnfthreeDTree_1$ and $n_2$ the number of side branches of the main branch of $\srnfthreeDTree_2$. We build a pairwise distance matrix $E$ of size $n_1 \times n_2$ as:
\begin{equation}
\label{eq:cost}
	E_{ij} = \lambda_p \left(\bifurcation^i_1 - \bifurcation_2^j\right)^2+\lambda_s \inf_{\greparm, \reparm} \| \srnf_1^i - (\srnf_2^j, \greparm,\reparm) \|^2, 
\end{equation}

\noi where $\srnf_1$ and $\srnf_2$ denote the side branches and $\bifurcation_1$ and $\bifurcation_2$ are the bifurcation points of those side branches to their parent branches, respectively. For the main branch, \ie the trunk of the entire tree, we set $\bifurcation_1$ and $\bifurcation_2$ to zeros. Then, we apply the Hungarian Algorithm to the matrix $E$ to find out the optimal matching $\permutation\optimal$ between the side branches and the associated optimum $\greparm\optimal$ and $\reparm\optimal$. At each layer of the trees, and for each branch, we build this matrix and recursively solve all problems from the highest layer to the main branch. Finally, we apply the optimal $\rotation\optimal$, $\greparm\optimal$, $\reparm\optimal$, and $\permutation\optimal$ to  $\threeDTree_2$ to get $\Bar{\threeDTree_2}$ registered on $\threeDTree_1$ as $\Bar{\threeDTree_2}=\rotation\optimal(\threeDTree_2,\greparm\optimal,\reparm\optimal,\permutation\optimal)$.

\begin{algorithm}[t]
\caption{Registration framework}
\label{alg:registration}
\raggedright
\SetKwInOut{Input}{Input}
\SetKwInOut{Output}{Output}
\SetKwFunction{AlignTree}{AlignTree} 
\SetKwProg{Fn}{Function}{:}{}
\Input{${L}$: number of layers of the input trees; $\srnfthreeDTree_1$ and  $\srnfthreeDTree_2$: the SRNF representation of two tree shapes;$(\lambda_m,\lambda_p,\lambda_s)$:  the controlling parameters of the metric.}
\Output{Optimal $\rotation\optimal, \greparm\optimal, \reparm\optimal, \permutation\optimal$  and distance cost metric $E$ of~\cref{eq:cost}.}
\textbf{Stage 1:} Find $\rotation\optimal$ of $\srnfthreeDTree_2$ that minimizes $d_{\preshapesrnf}$ [\cref{eq:metric_for_trees}] and set $\srnfthreeDTree_2 = \rotation\optimal(\srnfthreeDTree_2)$\;  
\textbf{Stage 2:} $E = AlignTree(\srnfthreeDTree_1, \srnfthreeDTree_2, L)$\; 
\textbf{Stage 3:} Output the optimal $\greparm\optimal$,$\reparm\optimal$,$\permutation\optimal$ for the combinations of branch pairs that yield minimum cost in $E$\; 
\Fn{$E \leftarrow \AlignTree(\srnfthreeDTree_1,\srnfthreeDTree_2,L)$}{
$E_p=$minimum distance for optimal $\greparm^0, \reparm^0$ that aligns $\srnf_1^0, \srnf_2^0$ [\cref{eq:cost}]\;

Initialize $E_c$ of size  $n_1\times n_2$  to $0$\;

\For {$i \in n_1$}{
\For{$j \in n_2$}{
$E_c(i,j)=$AlignTree($\srnfthreeDTree_1^i$, $\srnfthreeDTree_2^j$, $L-1$)\;
}
}
\Return $E=E_p+E_c$\;
}
\end{algorithm}

\section{Geodesics and summary statistics}
\label{sec:geod_stat}
In this section, we show how to use the building blocks developed in Sections~\ref{sec:representation} and~\ref{sec:elastic_metric} for computing geodesic paths between pairs of tree-shaped 3D objects, and summary statistics, such as means and modes of variation, of a collection of tree-shaped 3D objects. We perform the registration between 3D tree shapes $\preshapesrnf$ in the SRNF space. The geodesic between two tree shapes  can then be computed by taking a straight line (linear interpolation) between their SRNFs, and then mapping it back to the space of surfaces. This requires computing the inverse SRNF map. However, such an inverse SRNF map has no analytical formula. Laga \etal~\cite{laga2017numerical} developed a numerical solution that estimates  the inverse map. The approach, however,  is computationally very expensive, especially if used for complex tree-like 3D shapes that have many branch surfaces attached together. In this paper, we propose a practical solution to the inversion problem by introducing a new shape space based on tangents of 3D tree-like shapes, which is nearly equivalent to the SRNF representation, has an analytical inversion process for visualization, and is computationally more efficient.

\subsection{Geodesics}
\label{subsec:geod}
In this formulation, we represent a branch surface $\brachsurface\in\preshapebrachsurface$ with two orthogonal tangent vectors at each point of the surface as
\begin{equation}
   \label{eq:tan_space}
   \tanbranch(\brachsurface)= \left(\frac{F_u}{\sqrt{\norm{F_u}}},\frac{F_v}{\sqrt{\norm{F_v}}}\right) \text{ with } F_u=\frac{\partial\brachsurface}{\partial u} \text{ and } F_v=\frac{\partial\brachsurface}{\partial v}.
\end{equation}
\noi Here, $F_u$ and $F_v$  are the  tangent vectors to the surface in $u$ and $v$  directions.  

\vspace{3pt}
\noi\textbf{Equivalence of the representation to SRNF}: We rewrite~\cref{eq:srnf} as a cross product of two tangents that represent the normal $\surfacenormal$ at $(u,v)$ as:
\begin{equation}
    \label{eq:srnf_with_tan}
     \srnf(u,v)=\frac{F_u\times F_v}{\sqrt{\norm{F_u\times Fy}}} =\frac{\norm{F_u}\norm{F_v}\hat{n}}{\sqrt{\norm{Fx}\norm{F_v}\norm{\sin\theta}}},
\end{equation}

\noi where $\hat{n}$ is the unit normal at $(u,v)$, and $\theta$ is the angle between $F_u$ and $F_v$. Since branch shapes are analogous to the cylinder, which is our parameterization domain (\cref{subsec:elastic_metric_for_branch}), and we parameterize the branch as a smooth surface, the parameterization of the branch surface $\brachsurface$  is orthogonal with minimal error, \ie $\theta=90\degree$ and $\norm{\sin\theta}=1$. We can re-formulate~\cref{eq:srnf_with_tan} as: 
\begin{equation}
    \label{eq:eq_to_tan}
     \srnf(u,v)=\frac{\norm{F_u}\norm{F_v}\hat{n}}{\sqrt{\norm{F_u}\norm{F_v}}} = \sqrt{\norm{F_u}\norm{F_v}}\hat{n}.
\end{equation}

\noi  We thus can consider  $F_u$ and $F_v$ used in~\cref{eq:eq_to_tan} are the same as those in~\cref{eq:tan_space}, which are orthogonal to each other. Further, we can simplify~\cref{eq:tan_space} as:
\begin{equation}
   \label{eq:tan_space_simplified}
    \tangent(u,v)=\left(\sqrt{\norm{F_u}}a,\sqrt{\norm{F_v}b}\right), \text{ with } a=\frac{F_u}{\norm{F_u}}  \text{ and  } b=\frac{F_v}{\norm{F_v}}.
\end{equation}

\noi Our proposed representation quantifies the bending of the branches through the direction of tangents (\cref{eq:tan_space_simplified}), whereas in the SRNF representation, it is measured using the normal direction (\cref{eq:eq_to_tan}). Both SRNF and our tangent-based representation measure the amount of stretching through the length of two tangent vectors (see~\cref{eq:eq_to_tan} for SRNF and~\cref{eq:tan_space_simplified} for tangent-based formulation), which together encode the local surface area around a certain point. However, this equivalence does not hold for other shapes, \eg humans and animals, since orthogonal parameterization is not feasible for those shapes.

A complete 3D tree can now be represented using the same layering structure as defined in~\cref{sec:representation}, expressed as $\tangentTree^l = (\tangent^l,\{\tangentTree^{l+1}_i,s^{l}_i\}_{i=1}^{k^l}), \text{ for } l = 1, \dots, L-1$. Here, $L$ is the total number of layers,  $\tangent^l$ is the main branch of  $\tangentTree^l$, $k^l$ is the number of subtrees attached to the branch $\tangent^l$, and $\bifurcation^{l}_i \in [0, 1]$ is a bifurcation point on the medial axis of $\tangent^l$ of $\tangentTree^l$. We define the shape space of all such trees as $\preshapetan$. Note that, in $\preshapetan$, all shapes are registered, since we perform registration on the SRNF prior to mapping them into $\preshapetan$.

Since the tangent-based representation is equivalent to the SRNF, the $\ltwo$ metric in $\preshapetan$ is equivalent to the elastic metric. Thus,   geodesics under the complex metric become straight lines in the shape space $\preshapetan$, \ie we get the geodesic between two registered 3D trees $\tangentTree_1$ and $\tangentTree_2$ in $\preshapetan$ as linear interpolation: $\geod_\tangentTree(\tau)=(1-\tau)\tangentTree_1+\tau{\tangentTree}_2,\ \tau\in[0,1]$. For visualization, we map  $\geod_\tangentTree(\tau),\tau\in[0,1]$ back to the original tree shape space $\preshapespaceThreeDTreees$. In the inversion process, we consider the two components of~\cref{eq:tan_space} individually, solve them using the cumulative trapezoidal integration method~\cite{davis2014methods}, and then combine them by averaging to obtain the surface. 

\subsection{Statistics on 3D tree shapes}
\label{subsec:summary_stat}
In this section, we detail the computational tools used to compute various statistics of a collection of 3D tree shapes, including the mean and modes of shape variability. We obtain the mean, called the Karcher mean~\cite{wang2023elastic}, simultaneously when registering all shapes. However, since we perform registration in the SRNF space $\preshapesrnf$ and compute statistics in the tangent space $\preshapetan$, we compute two means in every iteration, one in SRNF space, $\Bar{\srnfthreeDTree}$ and one in tangent space, $\mu_\tangent$. Let we  $\{\threeDTree_1,\threeDTree_2,\dots,\threeDTree_n\}$ be  a collection of $n$ 3D tree shapes and  $\{\srnfthreeDTree_1,\srnfthreeDTree_2,\dots,\srnfthreeDTree_n\}$ their corresponding SRNF representations. We perform registration in the SRNF space to get all the shapes well aligned to each other, while computing the Karcher mean using the following optimization procedure:
\begin{enumerate}
	\item Set $\Bar{\srnfthreeDTree} =  \srnfthreeDTree_1$; 
    \item Map $\tangentTree_1\longleftarrow\threeDTree_1$; and set $ \mu_\tangent =  \tangentTree_1$. 
	\item \label{step:for2}for $i=1:n$
		\begin{itemize}
			\item Align  $\srnfthreeDTree_i$ onto $\Bar{\srnfthreeDTree}$ using~\cref{alg:registration}. 
			
                \item Let $\rotation\optimal_i$, $\greparm\optimal_i$,$\reparm\optimal_i$,$\permutation\optimal_i$be the optimums that align $\srnfthreeDTree_i$ onto $\Bar{\srnfthreeDTree}$.
                \item Update $\Tilde{\srnfthreeDTree_i} =  (\srnfthreeDTree_i,\rotation\optimal,\greparm\optimal,\reparm\optimal,\permutation\optimal)$.
                \item Update $\Bar{\srnfthreeDTree} =  \frac{1}{i}(\Bar{\srnfthreeDTree}+\Tilde{\srnfthreeDTree_i})$.
                \item Update $\Tilde{\threeDTree_i} = (\threeDTree_i,\rotation\optimal,\greparm\optimal,\reparm\optimal,\permutation\optimal)$.
                \item Map $\Tilde{\tangentTree}_i\longleftarrow\Tilde{\threeDTree}_i$
                \item Update $\mu_\tangent =  \frac{1}{i}(\mu_\tangent+\Tilde{\tangentTree_i})$.
		\end{itemize}	
	\item Return the set of registered 3D trees $\{\Tilde{\tangentTree_1},\Tilde{\tangentTree_2},\dots,\Tilde{\tangentTree_n}\}$. 
\end{enumerate}
The registered shapes $\{\Tilde{\tangentTree_i}\}_{i=1}^n$ are elements of  $\preshapetan$ equipped with the $\ltwo$ metric, thus one can perform  shape analysis tasks using the standard $\ltwo$ metric. 

\vspace{3pt}
\noi\textbf{Mean and modes of variation:} 
Given a set of registered 3D trees $\{\Tilde{\tangentTree_i}\}_{i=1}^n$ and their corresponding Karcher mean $\mu_\tangent$ in $\preshapetan$, we perform Principal Component Analysis (PCA) to get the statistics. Since the tangent-based representation space $\preshapetan$ is Euclidean, we compute the principal directions of variation using linear PCA. Let  $\covMatrix=\frac{1}{n-1}\sum_{i=1}^n(\Tilde{\tangentTree}^i-\mu_\tangent)(\Tilde{\tangentTree}^i-\mu_\tangent)^{\top}$ be the covariance matrix, $\{e_i\}_{i=1}^k$ its leading eigenvectors,  and $\{\delta_i\}_{i=1}^k$ the corresponding eigenvalues. The eigenvectors represent the principal directions of variation of the collection of trees $\{\Tilde{\tangentTree}_i\}_{i=1}^{n}$. The projection of any tree-shape onto the $i-$th principal direction is given by $\tangentTree_\tau=\mu_\tangent+\tau\sqrt{e_i}\delta_i,\ \tau\in\real$. For visualization, we map $\tangentTree_\tau$ back to the original space of 3D trees $\preshapespaceThreeDTreees$ using the inversion procedure described in~\cref{subsec:geod}. 

\vspace{3pt}
\noi\textbf{Generative model of 3D tree shapes:} We fit a multivariate Gaussian of mean $\mu_\tangent$ and diagonal covariance matrix $\covMatrix$ to characterize the shape variability in the collection of registered 3D tree shapes $\{\Tilde{\tangentTree}_i\}_{i=1}^{n}$. The diagonal elements of the covariance matrix $\covMatrix$ are the eigenvalues $\{\delta_i\}_{i=1}^k$.  
To generate a random 3D tree, we randomly sample $m$ real numbers $a_1,\dots,a_m\sim \mathcal{N}(0,1)$. A 3D tree-shape is then given by  $\tangentTree= \mu_\tangent + \sum_{i=1}^k a_i\sqrt{e}_i\delta_i$. We map $\tangentTree$
back to the original space using the inversion procedure described in~\cref{subsec:geod}. Note that, to ensure that the generated 3D tree-shapes are plausible, one can restrict the range of $a_i$'s, \eg to be within one standard deviation.

\section{Results and discussion}
\label{sec:results}
We evaluate the proposed framework using 3D tree models from~\cite{globeplant},~\cite{free3d},~\cite{wang2023elastic}, and real tomato plants from~\cite{pheno4d}. We first preprocess those models to convert them into our representation. The entire pipeline implemented in  MATLAB(2024), ran on a CPU with $2.00$GHz Intel(R) Core(TM) i$9$ processor and $128$GB of RAM. The supplementary material includes a detailed description of the dataset composition and the preprocessing mechanism.
\subsection{Representation}
\label{result_representation}

\begin{wrapfigure}{r}{0.7\linewidth}
    \centering
    \resizebox{\textwidth}{!}{
    \begin{tabular}{@{}ccccc@{}}
     \multicolumn{1}{c} {}&
    \multicolumn{3}{c} {
    \includegraphics[width=0.8\linewidth]{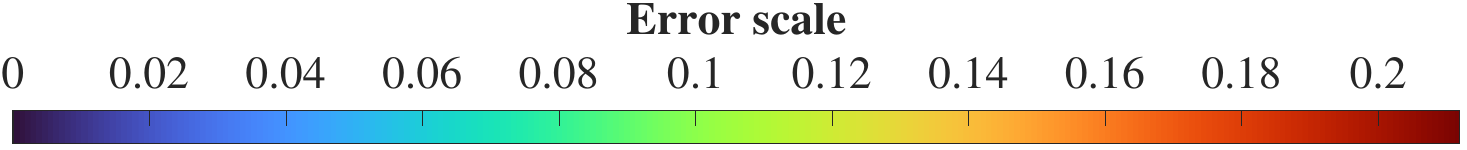}} & \multicolumn{1}{c} {}\\
     \includegraphics[width=0.2\linewidth]{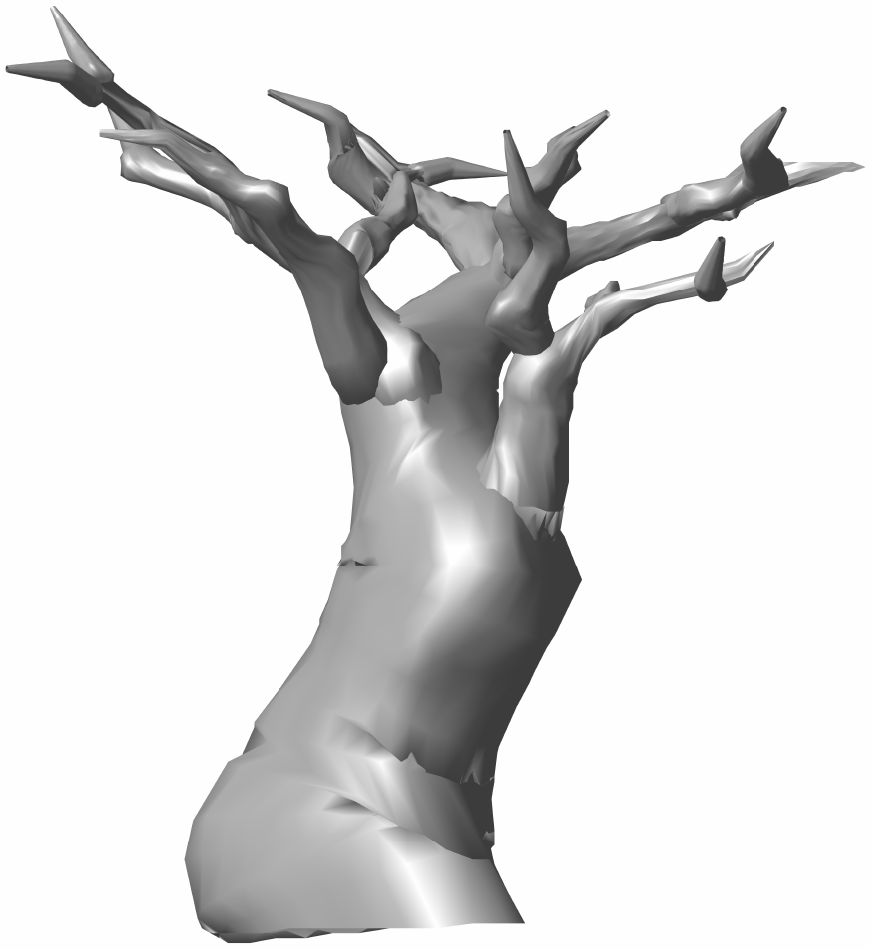} &  \includegraphics[width=0.2\linewidth]{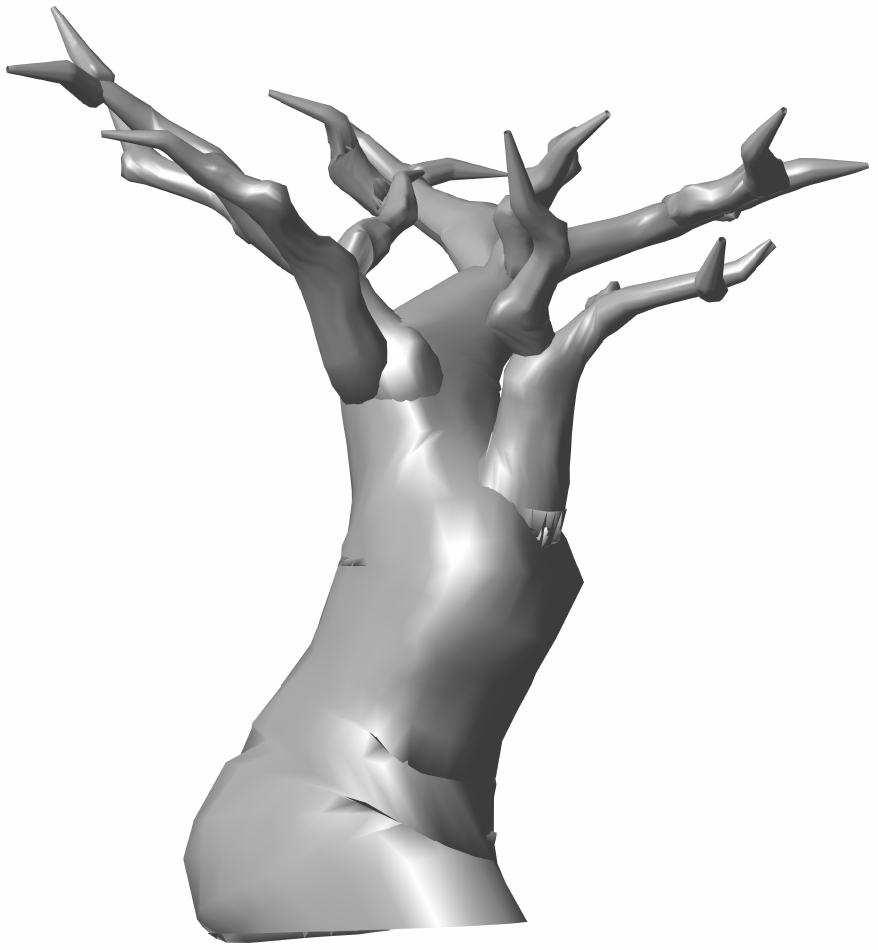} & \includegraphics[width=0.2\linewidth]{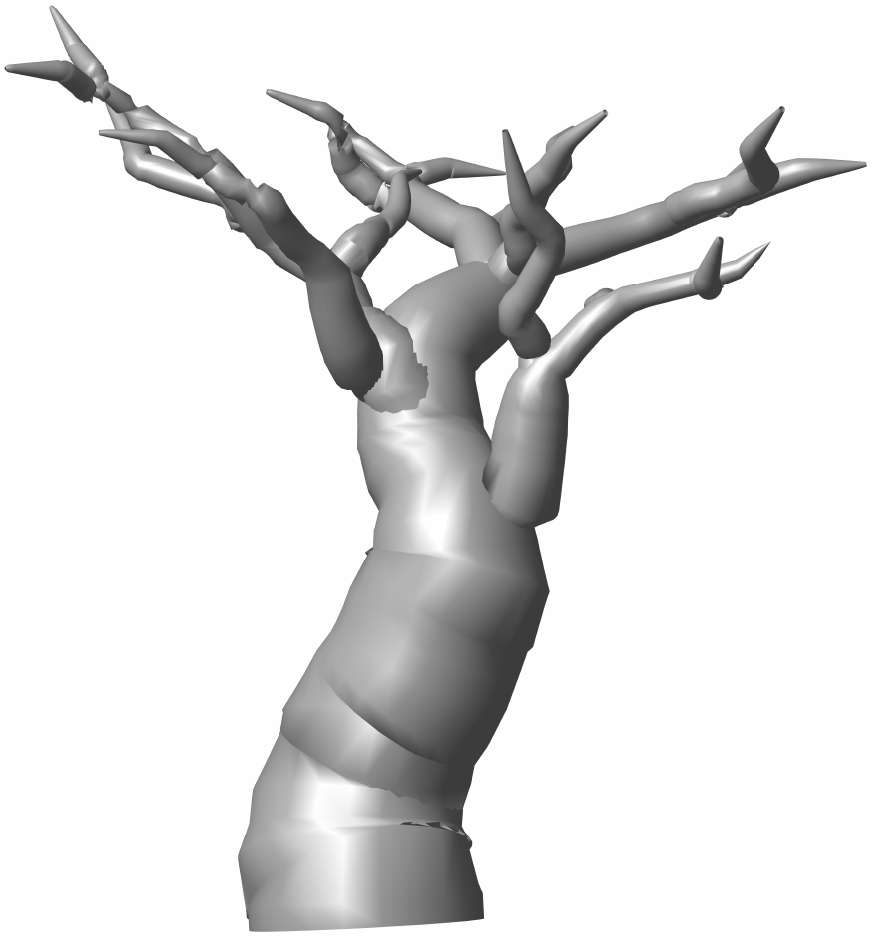} & \includegraphics[width=0.2\linewidth]{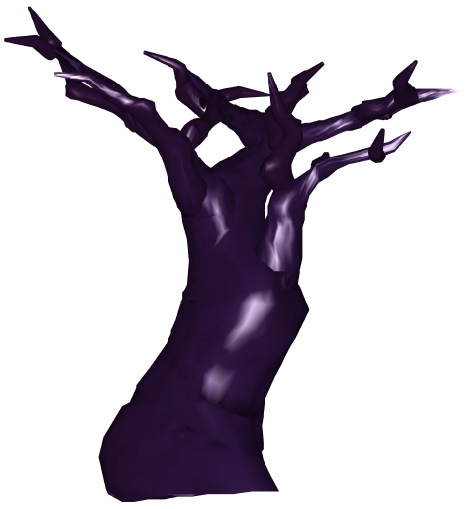} & \includegraphics[width=0.2\linewidth]{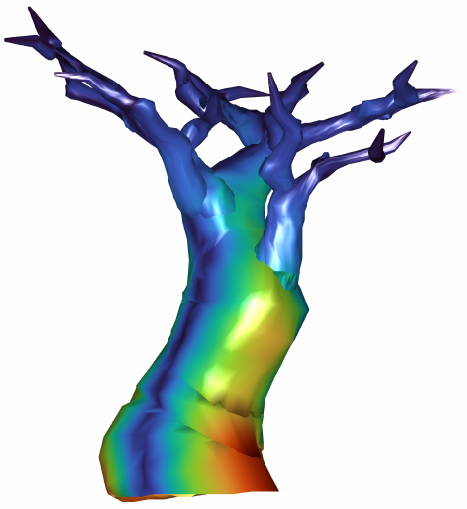} \\
      \small{\textbf{(a)} Original}   & \small{\textbf{(b)} Ours.} &  \small{\textbf{(c)}  Wang} & \small{\textbf{(d)} Pointwise} & \small{\textbf{(e)} Pointwise}\\ 
      \small{mesh.} &    & \small{\etal~\cite{wang2023elastic}'s.} & \small{error (ours).} & \small{error~\cite{wang2023elastic}.}
    \end{tabular}
    \caption{
    \label{fig:representation_qaulitative_quntitative}  Illustration of the accuracy of our representation.    }
    }
\end{wrapfigure} 
An important contribution of this paper is  that we represent the full geometry of tree-like 3D shapes, instead of approximating their branches with a skeleton augmented with thickness. 
~\cref{fig:representation_qaulitative_quntitative} compares visually the accuracy of this representation against previous methods. We can clearly see that the radial-based representation~ used in~\cite{wang2023elastic} (\cref{fig:representation_qaulitative_quntitative}-(c)) loses detailed surface information of the original 3D model (\cref{fig:representation_qaulitative_quntitative}-(a)) since it approximates the surface with a fixed radius circle. In contrast, our representation (\cref{fig:representation_qaulitative_quntitative}-(b)) accurately captures the surface details. To demonstrate this quantitatively, we compute the point-wise error between the original model and our surface-based representation, and the original surface and the radial-based representation~\cite{wang2023elastic}. We then plot this error in the form of a heatmap; see Figs.~\ref{fig:representation_qaulitative_quntitative}-(d) and (e), respectively. As one can see, the error in the radial-based representation is much higher than that of our surface-based representation, since~\cite {wang2023elastic} approximates the surface rather than representing it. The Supplementary Material includes more quantitative and qualitative results.

\subsection{Registration}
\label{result:registration}
\begin{figure}[t]
    \centering
    \begin{tabular}{@{}cccc@{}}
      \includegraphics[width=0.24\linewidth]{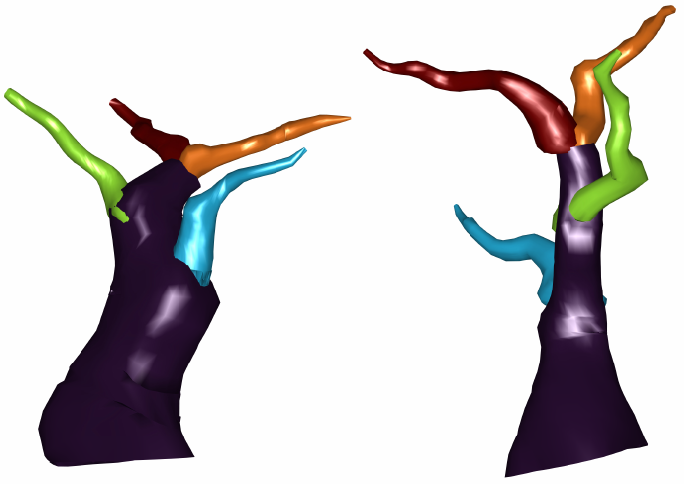} &
    \includegraphics[width=0.25\linewidth]{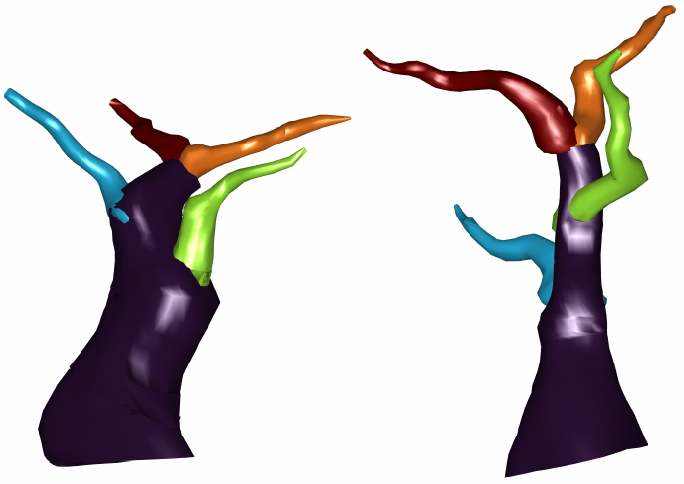} &
       \includegraphics[width=0.24\linewidth]{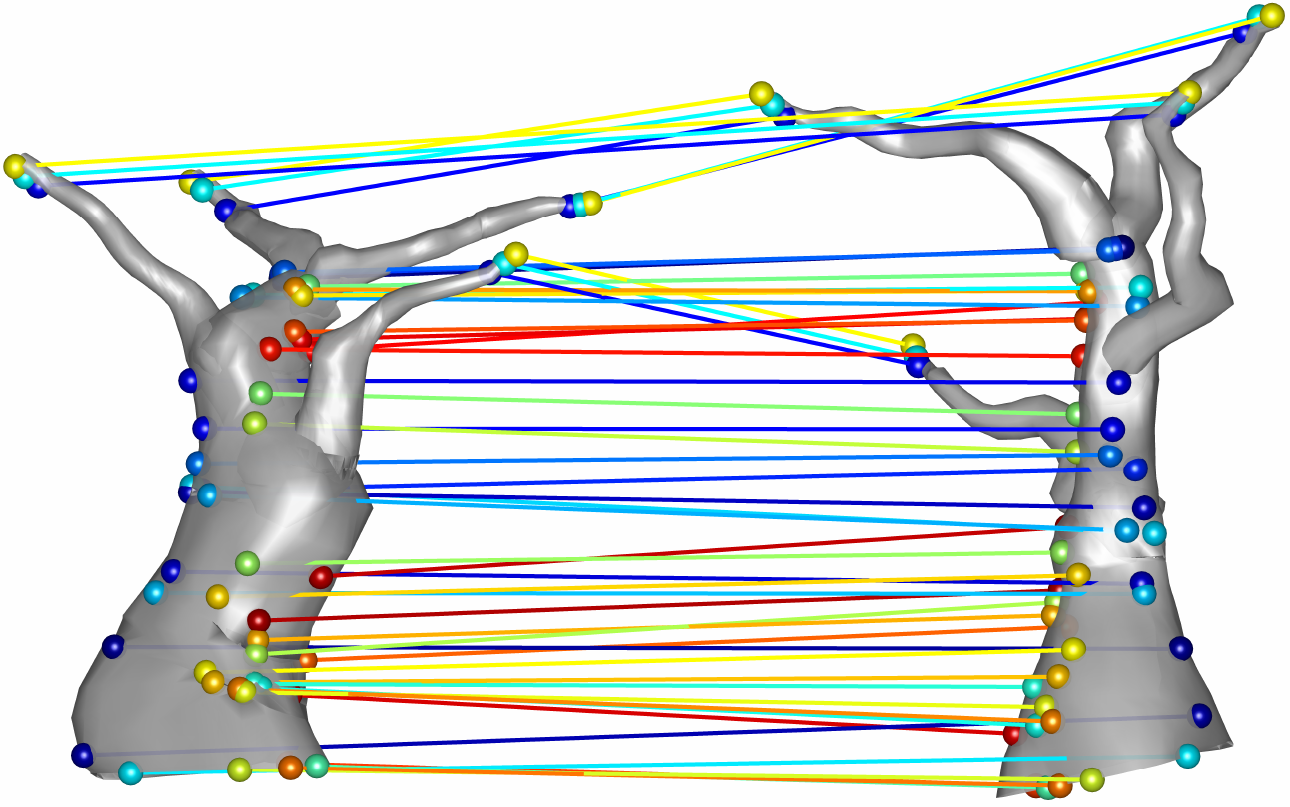} &
       \includegraphics[width=0.24\linewidth]{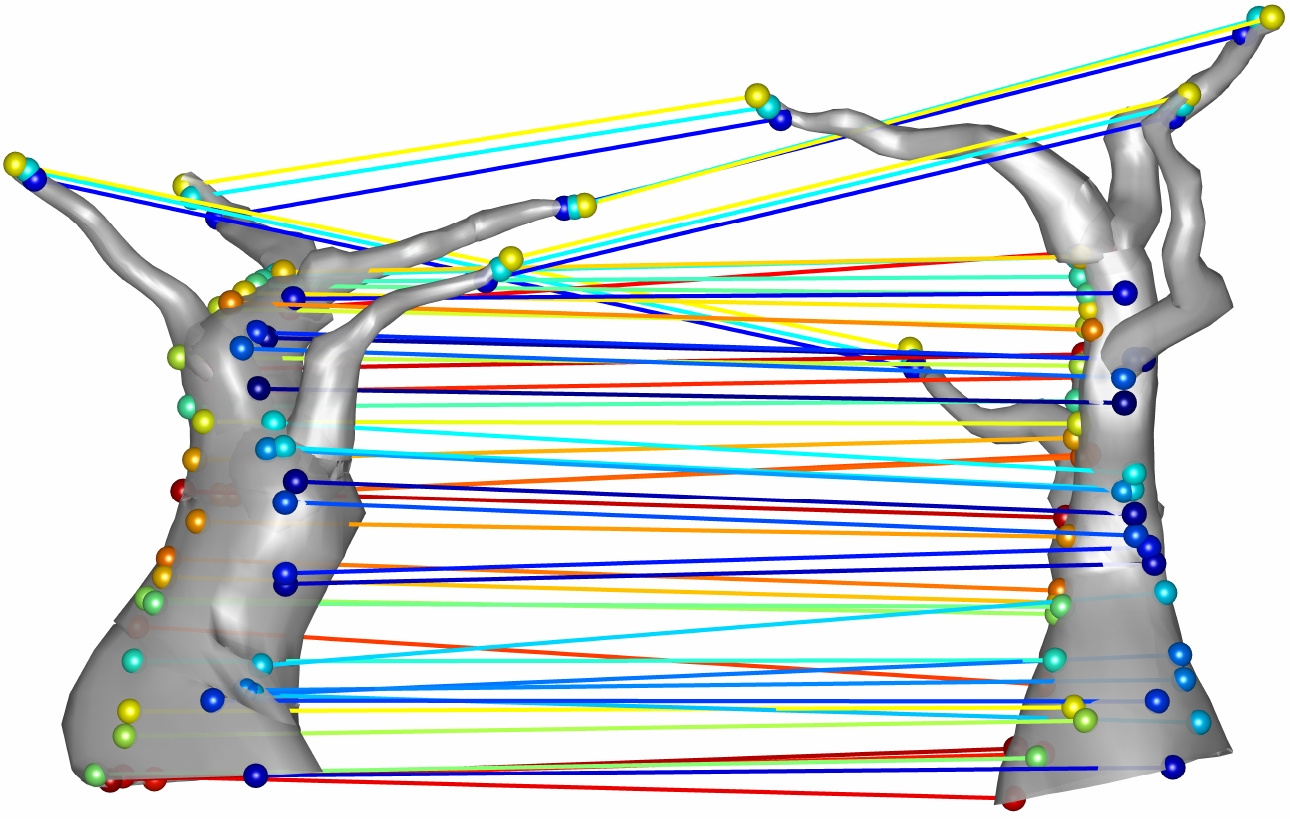}\\

       \small{\textbf{(a)} Before reg.} & 
       \small{\textbf{(b)} After reg.}  & 
        \small{\textbf{(c)} Before reg.}& 
        \small{\textbf{(d)} After reg.}\\
        
        \small{{(branchwise corres.)}} & 
         \small{{(branchwise corres.)}} &
        \small{{(pointwise corres.)}} &
        \small{{(pointwise corres.)}}
    \end{tabular}
    \caption{Registration result between two 3D botanical trees. We show branchwise correspondences \textbf{(a)} before and \textbf{(b)} after registration, and pointwise correspondences \textbf{(c)} before and \textbf{(d)} after registration. Correspondences are color-coded.}
    \label{fig:registration}
\end{figure}

\cref{fig:registration} shows an example of registration between two tree-shaped 3D objects obtained using our method. We show both the branch-wise and pointwise correspondences before and after applying our proposed algorithm.
For clarity, we show the correspondences up to the second layer of the hierarchy. We set the controlling parameters, $(\lambda_m,\lambda_p,\lambda_s)$, all to one to give equal importance to each term of the metric of~\cref{eq:tree_optim}. Fig. 8 in the Supplementary Material shows additional registration results between complex tree-shaped 3D objects. 
These results show that our registration method is able to find accurate branch-wise and point-wise correspondences, even between complex 3D trees with significant topological differences.
The Supplementary Material provides additional results, including an ablation study of the effects of varying the controlling parameters $(\lambda_m,\lambda_p,\lambda_s)$.

\begin{wraptable}{r}{0.70\textwidth} 
    \caption{Registration accuracy in terms of geodesic distance error $(\downarrow)$.}
    \begin{tabular}{|c|c|c|c|c|c|c|c|c|}
        \hline
             \multicolumn{3}{|c|}{\textbf{Before reg.}}  &  \multicolumn{3}{|c|}{\textbf{After reg. (ours)}} & \multicolumn{3}{|c|}{\textbf{After reg.~\cite{pan2021multi}}} \\
       \cline{1-9}
             Mean & Med. & Std. & Mean & Med. & Std. & Mean & Med. & Std.\\
        \hline
        \cline{1-9}
       $152.6$ & $179$ & $64$ & $\bf{75.3}$ & $\bf{77}$ & $\bf{29} $  &$140$ & $115$ & $95$\\
       \hline
    \end{tabular}
    \label{tab:geod_length}
\end{wraptable}

We quantitatively evaluate our proposed registration method and compare it with the state-of-the-art~\cite{pan2021multi} using four evaluation metrics: \textbf{(1)} the geodesic length between tree shapes, \textbf{(2)} cycle consistency,  \textbf{(3)} description length, and \textbf{(4)} computation time.  Note that the method of Wang \etal~\cite{wang2023elastic} follows the same pipeline as ours but only finds correspondences between tree skeletons. Ours, on the other hand, finds pointwise surface correspondences. Thus, these two methods cannot be compared.

\vspace{3pt}
\noi \textbf{(1) Geodesic length.}~\Cref{tab:geod_length} reports the mean, median, and standard deviation of the geodesic distances between sixteen pairs of 3D tree shapes before and after registration using our method and the method of~\cite{pan2021multi}. We can see that the geodesic distance decreases significantly after registration with our method, as the branches are correctly aligned across the 3D trees. The residual error is due to differences in the 3D tree structures. 

\begin{wraptable}{r}{0.6\textwidth}
    \caption{Cycle consistency errors $(\downarrow)$.}
    \label{tab:cycle_consistancy}
   \centering
    \begin{tabular}{|c|c|c|c|c|c|c|}
        \hline
            \multicolumn{1}{|c|}{} & \multicolumn{3}{|c|}{\textbf{Our method}}  &  \multicolumn{3}{|c|}{Pan \etal~\cite{pan2021multi}}\\
            \hline
        \cline{1-7}
          $\epsilon$   & Mean & Median & Std. & Mean & Median & Std.\\
        \hline
        \cline{1-7}
        $0.1$ & $\bf{0.1\%}$ & $\bf{0.1\%}$  & $\bf{0.08\%}$  & $20.5\%$ & $20\%$ & $7.6\%$\\
        \hline
        \cline{1-7}
        $0.05$ & $\bf{0.2\%}$  & $\bf{0.2\%}$  &  $\bf{0.11\%}$ & $29\%$ & $28\%$ & $5.2\%$\\  
        \hline
        \cline{1-7}
        $0.02$ & $\bf{1.6\%}$  &  $\bf{1.7\%}$ &  $\bf{0.97\%}$ & $32\%$ & $33\%$ & $2.1\%$ \\
        \hline
        \cline{1-7}
        $0.01$ & $\bf{3\%}$ & $\bf{2.9\%}$ &  $\bf{1.88\%}$ & $33.5\%$ & $33.2\%$  &  $1\%$ \\
        \hline
    \end{tabular}
\end{wraptable}
\vspace{3pt}
\noi \textbf{(2) Cycle consistency error}.
Given a source and a target 3D tree (sixteen random pairs considered), the registration process maps a point $\textbf{x}$ on the source to a point $\textbf{y}$ on the target. We then map $\textbf{y}$, using the same registration procedure, back onto the source, yielding a point $\textbf{x}'$. The registration procedure is accurate if  $\textbf{x}'$ and $\textbf{x}$ are very close to each other. Thus, we define the cycle consistency-based registration error as the percentage of points $\textbf{x}$ whose distance $\|\textbf{x} - \textbf{x}'\|$ is higher than a threshold $\epsilon$.  We take different values of $\epsilon$ between $0.01$ and $0.1$, and report in~\cref{tab:cycle_consistancy}  the mean, median, and standard deviation of this error for our method and~\cite{pan2021multi} (the lower, the better). As shown, our method significantly outperforms the state of the art.

\begin{wrapfigure}{r}{0.4\textwidth}
    \includegraphics[width=\linewidth]{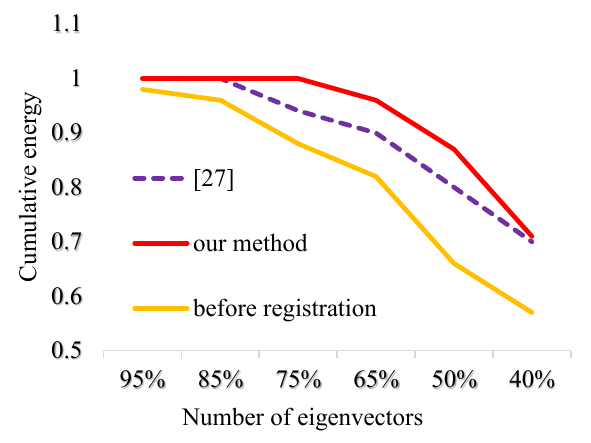}
    \caption{Cumulative energy with our method, with~\cite{pan2021multi} and on unregistered data.}
    \label{fig:description_length}
\end{wrapfigure}  

\vspace{3pt}
\noi\textbf{(3) Description length.} It is defined as the number of eigenvectors needed to describe $x\%$ of the variability within the dataset.  $x$ is called the cumulative energy. The higher the cumulative energy is, the better (see ~\cref{fig:description_length}).

\vspace{3pt}\noi\textbf{(4) Computation time.} Despite considering the entire surface of the 3D tree branches, our registration framework, on average, requires $43$ s to $20$ mins to align two 3D trees, whose number of branches ranges from $50$ to $15,000$. Pan \etal~\cite{pan2021multi}, on the other hand, requires $2.5$ mins to $6$ hrs for the same trees. 

\subsection{Geodesics}
\label{result:geod}
\begin{wrapfigure}{r}{0.65\linewidth}
    \centering
    \includegraphics[width=1\linewidth]{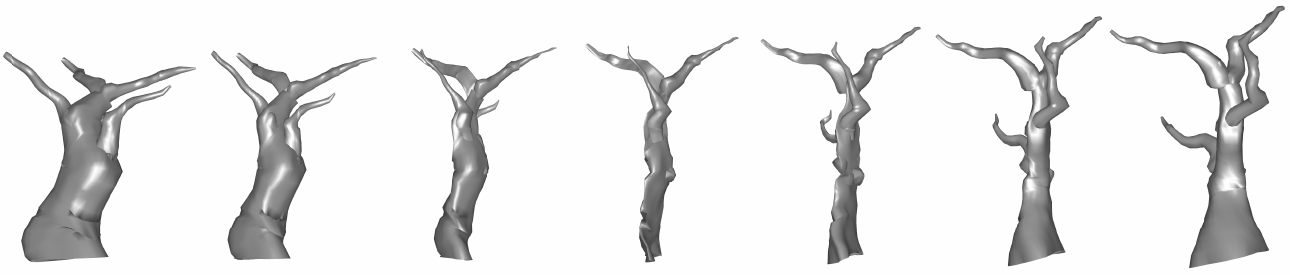}\\[-2pt]
    \small{\textbf{(a)} Geodesic before registration.}\\
    \includegraphics[width=1\linewidth]{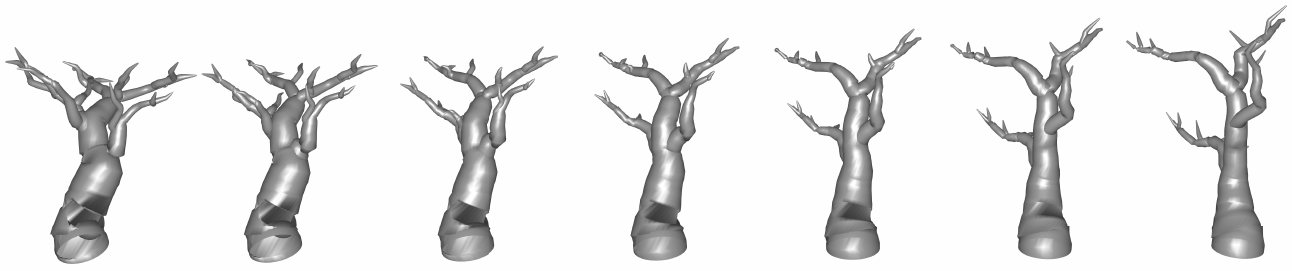}\\[-2pt]
    \small{\textbf{(b)} Geodesic using  Wang~\etal~\cite{wang2023elastic}.}\\
     \includegraphics[width=1\linewidth]{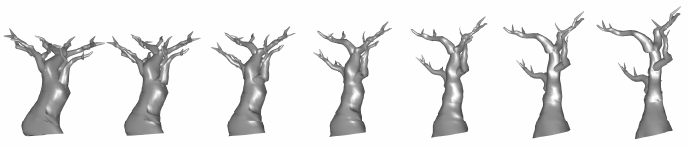}\\[-2pt]
    \small{\textbf{(c)} Geodesic using our method.}
    \caption{Geodesic between the leftmost (source) and the rightmost (target) 3D trees.   
   See also Fig. 15 in the Supplementary Material.}
    \label{fig:geod1}
\end{wrapfigure}

\cref{fig:geod1} shows a geodesic path between two 3D tree shapes, before registration (top row), with the approach of~\cite{wang2023elastic} (middle row), and with our approach (bottom row). We can clearly see that our proposed approach preserves rich geometry and topology along the geodesic path, whereas the path gets distorted before registration (top row), and approximation with radius in~\cite{wang2023elastic} (middle row) results in geometry loss. Our proposed approach works well on tree models with complex geometry and topology. The Supplementary Material provides more results.

\begin{wrapfigure}{r}{0.75\textwidth}
    \centering
    \includegraphics[trim=0 10 50 0, clip, width=\linewidth]{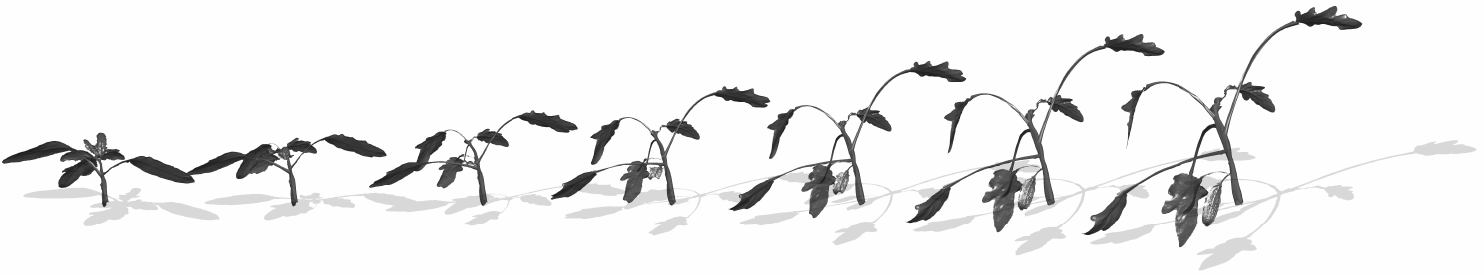} 
    \caption{Geodesic path between a small tomato plant~\cite{pheno4d} and a bigger one. This type of shape cannot be handled by previous methods such as~\cite{wang2023elastic}.}
    \label{fig:tomato_geod_growing_plant}
\end{wrapfigure}

We applied our registration and geodesics computation tools to 3D plants such as tomato plants whose leaves have complex geometries that cannot be modelled with previous techniques\cite{wang2023elastic}.~\cref{fig:tomato_geod_growing_plant} shows the geodesic between a small (at the very early growth stage) and a big (at a later growth stage) tomato plant. This confirms that our framework maintains its accuracy and robustness on noisy, organic, large-scale structures. 

\begin{wrapfigure}{r}{0.5\textwidth}
    \centering
    \includegraphics[trim=0 0 50 0, clip, width=\linewidth]{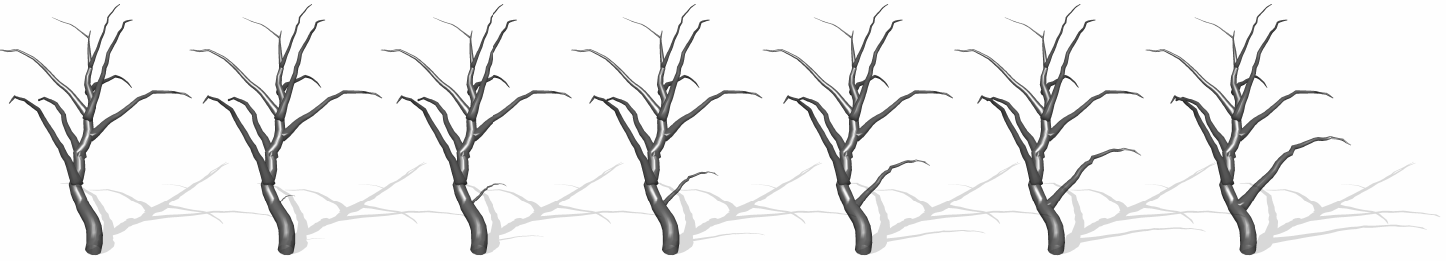}
    \caption{An example of geodesic between two 3D trees where the source (left tree) is generated by pruning the lower branch on the right side of the target (right tree). We can see that, using our method, the pruned branch naturally grows along the geodesic path.}
    \label{fig:branch_pruning}
\end{wrapfigure}

\cref{fig:branch_pruning} shows a geodesic path between a 3D tree and its version where one branch is pruned.  We can see that, when using our approach, the pruned branch naturally grows along the geodesic path. Fig. 51 in the supplementary material demonstrates the broader impact of our proposed tools by applying them to other datasets, such as neuronal structures. In terms of computation time, the geodesic computation takes, on average, $1.5$ s.

\subsection{Summary statistics}
\label{result:summary_stat}
\begin{wrapfigure}{r}{0.5\textwidth}
    \centering
    \begin{tabularx}{\textwidth}{YYYYYYY}
        $-1.5$ & $-1.0$ & $-0.5$ & $0$ & $+0.5$ & $+1.0$ & $+1.5$  \\[-3pt]
      \multicolumn{7}{c} {\includegraphics[width=1\linewidth]{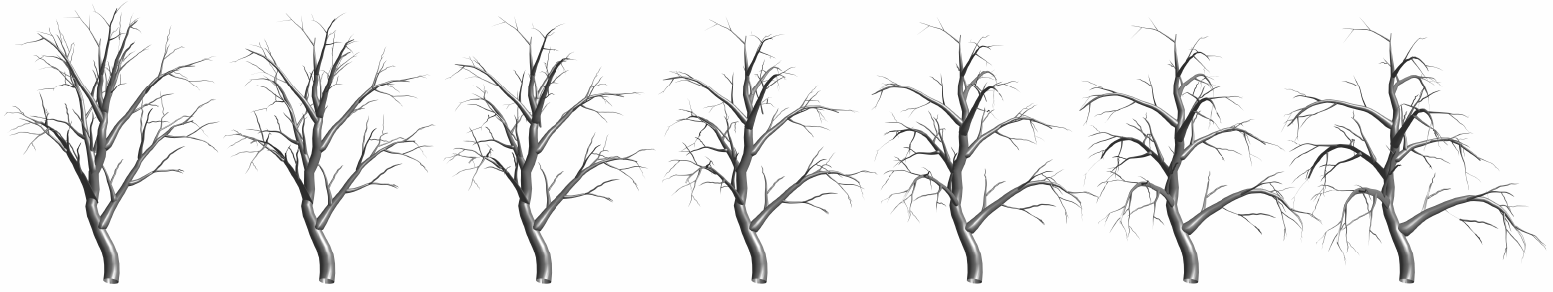}}\\[-3pt]
       \multicolumn{7}{c}
    {\begin{tikzpicture}
    \node[inner sep=0] (img) {\includegraphics[width=1\linewidth]{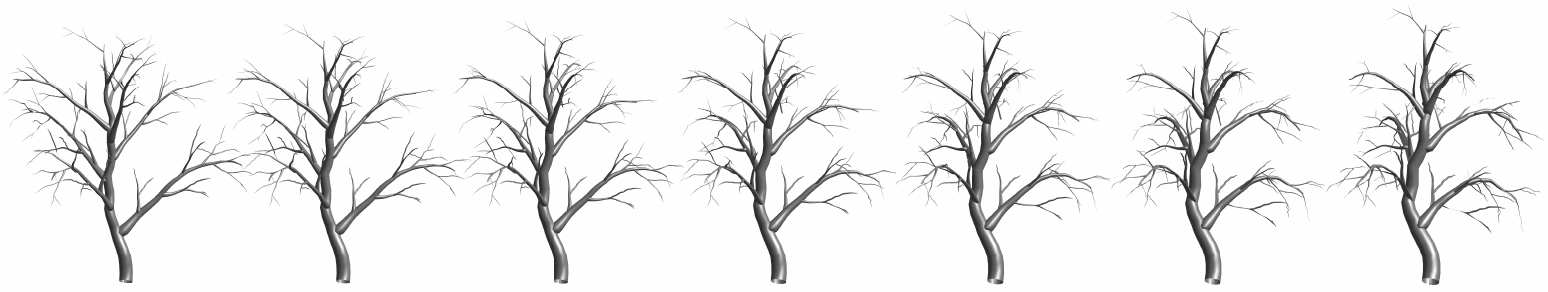}};
    \draw[red, thick] (-0.5,-0.7) rectangle (0.5,0.6);
     \end{tikzpicture}}\\[-3pt]
     \multicolumn{7}{c} {\includegraphics[width=1\linewidth]{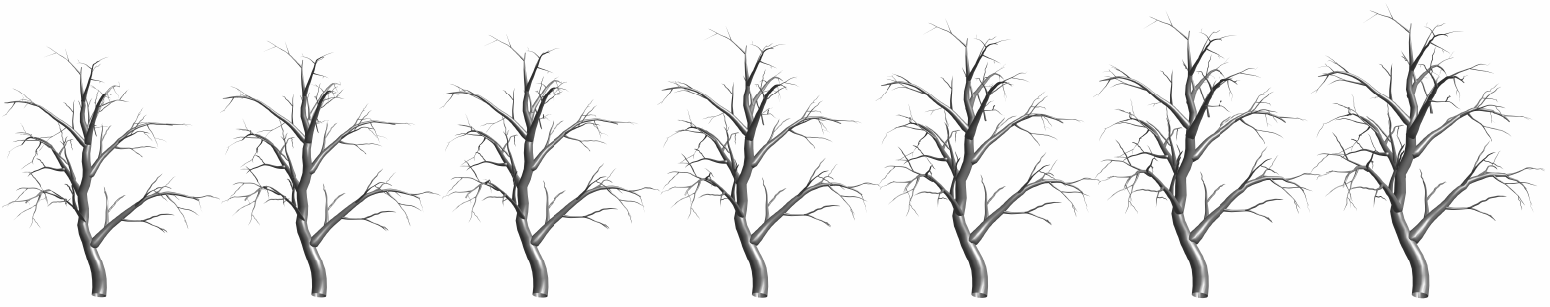}}
     \end{tabularx}
    \caption{The mean (in the red-colored box) and the first three leading principal modes of variation (one mode per row) 
    computed from the collection of 3D trees of Fig. 38 in the Supplementary Material.}
    \label{fig:mean_mode_set4_same_species}
\end{wrapfigure}

We group 3D trees in our dataset into several sets (see the Supplementary Material), each comprising unregistered trees of the same or different species. Then, we register and perform statistical analysis of the trees within each set.~\cref{fig:mean_mode_set4_same_species} shows the top three modes of variability (one per row) for the collection of trees from the same species.  The middle tree in each row corresponds to the mean of the collection. The Supplementary Material provides more results on other sets. We can see that the mean and leading principal directions capture the geometric and structural characteristics of the input collection. 

\begin{figure}[b]
    \centering
    \begin{tabular}{cc}
        \includegraphics[width=0.5\linewidth]{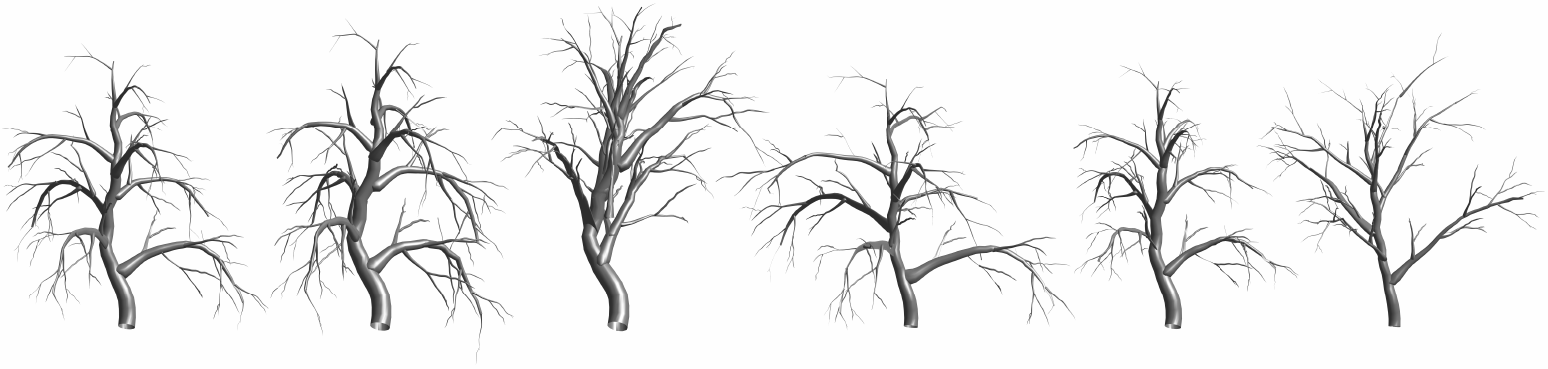} & \includegraphics[width=0.5\linewidth]{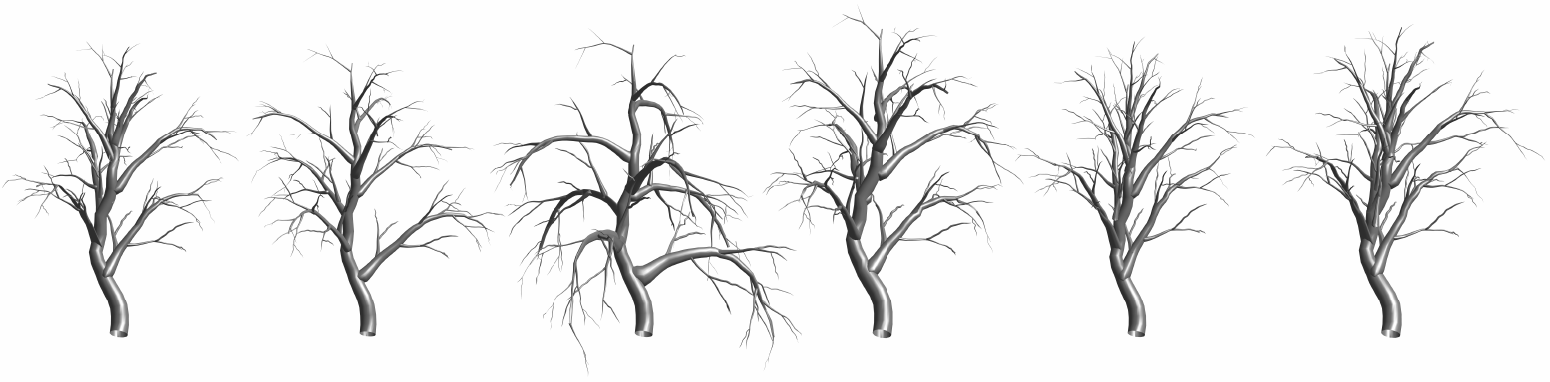}
    \end{tabular}  
    \caption{Examples of randomly generated 3D trees by sampling from the multivariate Gaussian fitted to the input 3D trees shown in Fig. 38 in the Supplementary Material.}
    \label{fig:rand_sample_from_set4}
\end{figure}

\vspace{3pt}
\noi \textbf{3D tree-shape generation.} To synthesize novel 3D tree shapes, we first fit a Gaussian distribution to a collection of  3D trees and then sample randomly from the distribution. We control the randomization so that the generation is plausible.~\cref{fig:rand_sample_from_set4} shows some examples of the generated trees. On average, the generation of one tree requires $0.01$ s.  We can see that the generated samples are unique while preserving the characteristics of the input trees. The Supplementary Material provides more results as well as a visualization of the 3D tree shapes used to learn the distributions. 

\section{Conclusion}
\label{sec:conclusion}
In this paper, we have offered computational tools to characterize the geometric and structural variability in tree-like 3D objects such as botanical trees and plants. Unlike previous works, we introduced a novel tree-shape space that encodes the entire 3D geometry in addition to structure. Using this novel tree-shape space, we have built computational tools for registration, geodesic computation,  statistical summarization, and generation of novel 3D tree shapes. Results demonstrate that our proposed framework outperforms the state-of-the-art.

While our current validation focuses on botanical trees and plants due to the availability of high-resolution 3D surface datasets, there are no theoretical constraints preventing its application to other tree-like structures, such as neurons or vascular networks. By demonstrating robustness on botanical trees, which exhibit extreme geometric and topological variation, we provide strong empirical evidence for the framework’s broader applicability. The supplementary material shows some results on neuronal structures. As future work, we plan to extend the shape space and the computational tools to modelling the spatiotemporal variability in 4D tree-like shapes.

\vspace{3pt}
\noi\textbf{Acknowledgment.} This work is supported by the Australian Research Council (ARC) Discovery Projects no. DP210101682, DP220102197,  and DP260101891, and ARC Future Fellowship FT250100448. Tahmina Khanam is funded by the Murdoch University Postgraduate Scholarship.

%
%
\clearpage
\newpage
\bibliographystyle{splncs04}
\bibliography{main}
\end{document}

%% file: notation_3Dtree.tex
\newcommand\norm[1]{\left\lVert#1\right\rVert}

\newcommand{\branch}{\beta}
\newcommand{\medialaxis}{f}
\newcommand{\threeDTree}{T}
\newcommand{\brachsurface}{F}

\newcommand{\paramdomain}{\Omega}
\newcommand{\parameters}{\omega}

\newcommand{\distance}{d}
\newcommand{\distancebranch}{d_F}

\newcommand{\srnf}{q}
\newcommand{\srnfbranch}{H}
\newcommand{\srnfthreeDTree}{Q}
\newcommand{\surfacenormal}{n}
\newcommand{\tanbranch}{G}
\newcommand{\tangent}{w}
\newcommand{\tangentTree}{W}

\newcommand{\preshapespace}{\mathcal{C}}
\newcommand{\preshapebrachsurface}{\preshapespace_\brachsurface}
\newcommand{\preshapespaceThreeDTreees}{\preshapespace_\threeDTree}
\newcommand{\preshapespacesrnfbranch}{\preshapespace_\srnf}
\newcommand{\preshapesrnf}{\preshapespace_\srnfthreeDTree}
\newcommand{\preshapetan}{\preshapespace_\tangentTree}
\newcommand{\rotation}{O}
\newcommand{\rotationspace}{SO(3)}
\newcommand{\reparm}{\gamma}
\newcommand{\greparm}{\epsilon}
\newcommand{\reparmspace}{\Gamma}

\newcommand{\permutation}{\sigma}
\newcommand{\permutationspace}{S}

\newcommand{\optimal}{^*}

\def\argmin{\mathop{\rm argmin}}

\newcommand{\ltwo}{\mathbb{L}^2}
\newcommand{\real}{\mathbb{R}}
\newcommand{\rthree}{\real^3}
\newcommand{\sone}{S^1}

\newcommand{\unitcylinder}{[0,1]\times \sone}

\newcommand{\noi}{\noindent}
\newcommand{\bifurcation}{s}

\newcommand{\covMatrix}{K}

\newcommand{\geod}{\Lambda}

%% file: main.bib
@String(IJCV  = {Int. J. Comput. Vis.})

@String(CVPR  = {IEEE Conf. Comput. Vis. Pattern Recog.})

@String(ICCV  = {Int. Conf. Comput. Vis.})

@String(ECCV  = {Eur. Conf. Comput. Vis.})

@String(ACCV  = {Asian Conf. Comput. Vis.})

@String(TOG   = {ACM Trans. Graph.})

@String(VR    = {Vis. Res.})

@String(IJCV  = {IJCV})

@String(CVPR  = {CVPR})

@String(ICCV  = {ICCV})

@String(ECCV  = {ECCV})

@String(ACCV  = {ACCV})

@String(TOG   = {ACM TOG})

@article{wang2023elastic,
  title={Elastic shape analysis of tree-like 3d objects using extended srvf representation},
  author={Wang, Guan and Laga, Hamid and Srivastava, Anuj},
  journal={IEEE transactions on pattern analysis and machine intelligence},
  volume={46},
  number={4},
  pages={2475--2488},
  year={2023},
  publisher={IEEE},
  doi={10.1109/TPAMI.2023.3334525}
}

@inproceedings{jermyn2012elastic,
  title={Elastic shape matching of parameterized surfaces using square root normal fields},
  author={Jermyn, Ian H and Kurtek, Sebastian and Klassen, Eric and Srivastava, Anuj},
  booktitle={ECCV},
  pages={804--817},
  year={2012},
  organization={Springer},
  doi={10.1007/978-3-642-33715-4_58}
}

@book{jermyn2017elastic,
  title={Elastic shape analysis of three-dimensional objects},
  author={Jermyn, Ian H and Kurtek, Sebastian and Laga, Hamid and Srivastava, Anuj and Medioni, Gerard and Dickinson, Sven},
  volume={12},
  year={2017},
  publisher={Springer},
  doi={10.1007/978-3-031-01819-0}
}

@article{duncan2018statistical,
  title={Statistical shape analysis of simplified neuronal trees},
  author={Duncan, Adam and Klassen, Eric and Srivastava, Anuj},
  journal={The Annals of Applied Statistics},
  volume={12},
  number={3},
  pages={1385--1421},
  year={2018},
  publisher={JSTOR},
  url={https://www.jstor.org/stable/26542579}
}

@article{laga2017numerical,
  title={Numerical inversion of SRNF maps for elastic shape analysis of genus-zero surfaces},
  author={Laga, Hamid and Xie, Qian and Jermyn, Ian H and Srivastava, Anuj and others},
  journal={IEEE transactions on pattern analysis and machine intelligence},
  volume={39},
  number={12},
  pages={2451--2464},
  year={2017},
  publisher={IEEE},
  doi={10.1109/TPAMI.2016.2647596}
}

@inproceedings{laga2017modeling,
  title={Modeling and Exploring Co-variations in the Geometry and Configuration of Man-made 3D Shape Families},
  author={Laga, Hamid and Tabia, Hedi},
  booktitle={Computer Graphics Forum},
  volume={36},
  number={5},
  pages={13--25},
  year={2017},
  organization={Wiley Online Library},
  doi={10.1111/cgf.13241}
}

@article{laga20224d,
  title={4D atlas: Statistical analysis of the spatiotemporal variability in longitudinal 3D shape data},
  author={Laga, Hamid and Padilla, Marcel and Jermyn, Ian H and Kurtek, Sebastian and Bennamoun, Mohammed and Srivastava, Anuj},
  journal={IEEE TPAMI},
  volume={45},
  number={2},
  pages={1335--1352},
  year={2022},
  publisher={IEEE},
  doi={10.1109/TPAMI.2022.3163720}
}

@inproceedings{feragen2010geometries,
  title={Geometries on spaces of treelike shapes},
  author={Feragen, Aasa and Lauze, Francois and Lo, Pechin and de Bruijne, Marleen and Nielsen, Mads},
  booktitle={ACCV},
  pages={160--173},
  year={2010},
  organization={Springer},
  doi={10.1007/978-3-642-19309-5_13}
}

@inproceedings{feragen2011means,
  title={Means in spaces of tree-like shapes},
  author={Feragen, Aasa and Hauberg, S{\o}ren and Nielsen, Mads and Lauze, Fran{\c{c}}ois},
  booktitle={ICCV},
  pages={736--746},
  year={2011},
  organization={IEEE},
  doi={10.1109/ICCV.2011.6126311}
}

@article{feragen2012toward,
  title={Toward a theory of statistical tree-shape analysis},
  author={Feragen, Aasa and Lo, Pechin and de Bruijne, Marleen and Nielsen, Mads and Lauze, Fran{\c{c}}ois},
  journal={IEEE TPAMI},
  volume={35},
  number={8},
  pages={2008--2021},
  year={2012},
  publisher={IEEE},
  doi={10.1109/TPAMI.2012.265}
}

@inproceedings{feragen2013tree,
  title={Tree-space statistics and approximations for large-scale analysis of anatomical trees},
  author={Feragen, Aasa and Owen, Megan and Petersen, Jens and Wille, Mathilde MW and Thomsen, Laura H and Dirksen, Asger and de Bruijne, Marleen},
  booktitle={Information Processing in Medical Imaging: 23rd International Conference},
  pages={74--85},
  year={2013},
  organization={Springer},
  doi={10.1007/978-3-642-38868-2_7}
}

@article{wang2018shape,
  title={The shape space of 3D botanical tree models},
  author={Wang, Guan and Laga, Hamid and Xie, Ning and Jia, Jinyuan and Tabia, Hedi},
  journal={ACM ToG},
  volume={37},
  number={1},
  pages={1--18},
  year={2018},
  publisher={ACM New York, NY, USA},
  doi={10.1145/3144456}
}

@inproceedings{wang2018statistical,
  title={Statistical modeling of the 3D geometry and topology of botanical trees},
  author={Wang, Guan and Laga, Hamid and Jia, Jinyuan and Xie, Ning and Tabia, Hedi},
  booktitle={Computer graphics forum},
  volume={37},
  number={5},
  pages={185--198},
  year={2018},
  doi={10.1111/cgf.13501}
}

@incollection{kilian2007geometric,
  title={Geometric modeling in shape space},
  author={Kilian, Martin and Mitra, Niloy J and Pottmann, Helmut},
  booktitle={ACM SIGGRAPH},
  pages={64--es},
  year={2007},
  doi={10.1145/1275808.1276457}
}

@article{wirth2011continuum,
  title={A continuum mechanical approach to geodesics in shape space},
  author={Wirth, Benedikt and Bar, Leah and Rumpf, Martin and Sapiro, Guillermo},
  journal={IJCV},
  volume={93},
  pages={293--318},
  year={2011},
  publisher={Springer},
  doi={10.1007/s11263-010-0416-9}
}

@inproceedings{zhang2015shell,
  title={{Shell PCA: statistical shape modelling in shell space}},
  author={Zhang, Chao and Heeren, Behrend and Rumpf, Martin and Smith, William AP},
  booktitle={Proceedings of the IEEE ICCV},
  pages={1671--1679},
  year={2015},
  doi={10.1109/ICCV.2015.195}
}

@article{laga2018survey,
  title={A survey on nonrigid 3d shape analysis},
  author={Laga, Hamid},
  journal={Academic Press Library in Signal Processing, Volume 6},
  pages={261--304},
  year={2018},
  publisher={Elsevier},
  doi={10.1016/B978-0-12-811889-4.00007-5}
}

@inproceedings{heeren2012time,
  title={Time-discrete geodesics in the space of shells},
  author={Heeren, Behrend and Rumpf, Martin and Wardetzky, Max and Wirth, Benedikt},
  booktitle={Computer Graphics Forum},
  volume={31},
  number={5},
  pages={1755--1764},
  year={2012},
  doi={10.1111/j.1467-8659.2012.03180.x}
}

@article{billera2001geometry,
  title={Geometry of the space of phylogenetic trees},
  author={Billera, Louis J and Holmes, Susan P and Vogtmann, Karen},
  journal={Advances in Applied Mathematics},
  volume={27},
  number={4},
  pages={733--767},
  year={2001},
  publisher={Elsevier},
  doi={10.1006/aama.2001.0759}
}

@article{owen2010fast,
  title={A fast algorithm for computing geodesic distances in tree space},
  author={Owen, Megan and Provan, J Scott},
  journal={IEEE/ACM Transactions on Computational Biology and Bioinformatics},
  volume={8},
  number={1},
  pages={2--13},
  year={2010},
  publisher={IEEE},
  doi={10.1109/TCBB.2010.3}
}

@article{zhang2023spatio,
  title={Spatio-temporal registration of plants non-rigid 3-D structure},
  author={Zhang, Tian and Elnashef, Bashar and Filin, Sagi},
  journal={ISPRS Journal of Photogrammetry and Remote Sensing},
  volume={205},
  pages={263--283},
  year={2023},
  publisher={Elsevier},
  doi={10.1016/j.isprsjprs.2023.10.009}
}

@article{wang2022plantmove,
  title={PlantMove: A tool for quantifying motion fields of plant movements from point cloud time series},
  author={Wang, Di and Puttonen, Eetu and Casella, Eric},
  journal={International Journal of Applied Earth Observation and Geoinformation},
  volume={110},
  pages={102781},
  year={2022},
  publisher={Elsevier},
  doi={https://doi.org/10.1016/j.jag.2022.102781}
}

@inproceedings{lobefaro2023estimating,
  title={Estimating 4D Data Associations Towards Spatial-Temporal Mapping of Growing Plants for Agricultural Robots},
  author={Lobefaro, Luca and Malladi, Meher VR and Vysotska, Olga and Guadagnino, Tiziano and Stachniss, Cyrill},
  booktitle={IROS},
  pages={4212--4218},
  year={2023},
  organization={IEEE},
  doi={10.1109/IROS55552.2023.10342449}
}

@inproceedings{magistri2020segmentation,
  title={Segmentation-based 4D registration of plants point clouds for phenotyping},
  author={Magistri, Federico and Chebrolu, Nived and Stachniss, Cyrill},
  booktitle={IROS},
  pages={2433--2439},
  year={2020},
  organization={IEEE},
  doi={10.1109/IROS45743.2020.9340918}
}

@inproceedings{pan2021multi,
  title={Multi-scale space-time registration of growing plants},
  author={Pan, Haolin and H{\'e}troy-Wheeler, Franck and Charlaix, Julie and Colliaux, David},
  booktitle={2021 International Conference on 3D Vision (3DV)},
  pages={310--319},
  year={2021},
  organization={IEEE},
  doi={10.1109/3DV53792.2021.00041}
}

@article{chebrolu2021registration,
  title={Registration of spatio-temporal point clouds of plants for phenotyping},
  author={Chebrolu, Nived and Magistri, Federico and L{\"a}be, Thomas and Stachniss, Cyrill},
  journal={PloS one},
  volume={16},
  number={2},
  pages={e0247243},
  year={2021},
  publisher={Public Library of Science San Francisco, CA USA},
  doi={10.1371/journal.pone.0247243}
}

@inproceedings{khanam2024riemannian,
  title={A Riemannian Approach for Spatiotemporal Analysis and Generation of 4D Tree-shaped Structures},
  author={Khanam, Tahmina and Laga, Hamid and Bennamoun, Mohammed and Wang, Guanjin and Sohel, Ferdous and Boussaid, Farid and Wang, Guan and Srivastava, Anuj},
  booktitle={ECCV},
  pages={326--341},
  year={2024},
  organization={Springer},
  doi={10.1007/978-3-031-72855-6_19}
}

@inproceedings{wang2025autoregressive,
  title={Autoregressive generation of static and growing trees},
  author={Wang, Hanxiao and Zhang, Biao and Klein, Jonathan and Michels, Dominik L and Yan, Dong-Ming and Wonka, Peter},
  booktitle={Proceedings of the SIGGRAPH Asia 2025 Conference Papers},
  pages={1--12},
  year={2025},
  doi={10.1145/3757377.3763818}
}

@inproceedings{lee2024tree,
  title={Tree-D Fusion: Simulation-Ready Tree Dataset from Single Images with Diffusion Priors},
  author={Lee, Jae Joong and Li, Bosheng and Beery, Sara and Huang, Jonathan and Fei, Songlin and Yeh, Raymond A and Benes, Bedrich},
  booktitle={ECCV},
  pages={439--460},
  year={2024},
  organization={Springer},
  doi={10.1007/978-3-031-72940-9_25}
}

@article{zhou2023deeptree,
  title={Deeptree: Modeling trees with situated latents},
  author={Zhou, Xiaochen and Li, Bosheng and Benes, Bedrich and Fei, Songlin and Pirk, S{\"o}ren},
  journal={IEEE Transactions on Visualization and Computer Graphics},
  volume={30},
  number={8},
  pages={5795--5809},
  year={2023},
  publisher={IEEE},
  doi={10.1109/TVCG.2023.3307887}
}

@article{li2024interactive,
  title={Interactive invigoration: Volumetric modeling of trees with strands},
  author={Li, Bosheng and Schwarz, Nikolas Alexander and Pa{\l}ubicki, Wojtek and Pirk, S{\"o}ren and Benes, Bedrich},
  journal={ACM Transactions on Graphics (TOG)},
  volume={43},
  number={4},
  pages={1--13},
  year={2024},
  publisher={ACM New York, NY, USA},
  doi={10.1145/3658206}
}

@ARTICLE{zhou2025treestructor,
  author={Zhou, Xiaochen and Li, Bosheng and Benes, Bedrich and Habib, Ayman and Fei, Songlin and Shao, Jinyuan and Pirk, Sören},
  journal={IEEE Transactions on Geoscience and Remote Sensing}, 
  title={TreeStructor: Forest Reconstruction With Neural Ranking}, 
  year={2025},
  volume={63},
  number={},
  pages={1-19},
  doi={10.1109/TGRS.2025.3558312}}

@inproceedings{li2025stressful,
  title={Stressful Tree Modeling: Breaking Branches with Strands},
  author={Li, Bosheng and Schwarz, Nikolas and Palubicki, Wojtek and Pirk, S{\"o}ren and Michels, Dominik L and Benes, Bedrich},
  booktitle={Proceedings of the Special Interest Group on Computer Graphics and Interactive Techniques Conference Conference Papers},
  pages={1--11},
  year={2025},
  doi={10.1145/3721238.3730745}
}

@InProceedings{Liu_2025_CVPR,
    author    = {Liu, Zhihao and Cheng, Zhanglin and Yokoya, Naoto},
    title     = {Neural Hierarchical Decomposition for Single Image Plant Modeling},
    booktitle = {Proceedings of the IEEE/CVF Conference on Computer Vision and Pattern Recognition (CVPR)},
    month     = {June},
    year      = {2025},
    pages     = {733-742},
    doi={10.1109/CVPR52734.2025.00077}
}

@article{pheno4d,
  title={Pheno4D: A spatio-temporal dataset of maize and tomato plant point clouds for phenotyping and advanced plant analysis},
  author={Schunck, David and Magistri, Federico and Rosu, Radu Alexandru and Corneli{\ss}en, Andr{\'e} and Chebrolu, Nived and Paulus, Stefan and L{\'e}on, Jens and Behnke, Sven and Stachniss, Cyrill and Kuhlmann, Heiner and others},
  journal={Plos one},
  volume={16},
  number={8},
  pages={e0256340},
  year={2021},
  publisher={Public Library of Science San Francisco, CA USA},
  doi={10.1371/journal.pone.0256340}
}

@book{davis2014methods,
  title={Methods of Numerical Integration},
  author={Davis, P.J. and Rabinowitz, P. and Rheinbolt, W.},
  isbn={9781483264288},
  series={Computer Science and Applied Mathematics},
  url={https://books.google.com.au/books?id=mbLiBQAAQBAJ},
  year={2014},
  publisher={Academic Press}
}

@misc{globeplant,
  author       = {},
  title        = {Globe Plants},
  year         = {},
  url          = {https://globeplants.com},
  note         = {Accessed: 24 June 2026.},
  urldate  = {2026-06-24}
}

@misc{free3d,
  author       = {},
  title        = {Free3DModels},
  year         = {},
  url          = {https://free3d.com/},
  note         = {Accessed: 24 June 2026},
  urldate  = {2026-06-24}
}
